%% file: soft-grasp-arxiv2016.tex
\documentclass[a4paper, 10pt, conference]{ieeeconf}

\IEEEoverridecommandlockouts

\usepackage{amssymb,url,amsmath}
\usepackage{algorithm}
\usepackage{graphicx}
\usepackage{epstopdf}
\usepackage{textcomp}
\usepackage{color}
\usepackage{ifpdf}
\usepackage{url}
\usepackage{enumerate}
\usepackage{array}
\usepackage{flushend}
\usepackage{multirow,booktabs}
\usepackage{hyperref}
\usepackage{subfig}
\usepackage{mdwlist}
\usepackage[utf8]{inputenc}
\usepackage{tcolorbox}
\newcommand{\alg}{Algorithm~}
\newtcolorbox[auto counter]{algorithmbox}[2][]{colback=red!5!white,colframe=red!75!black,fonttitle=\bfseries, title=\alg\thetcbcounter: #2,#1}

\newcommand{\eq}{Eq.~}
\newcommand{\fig}{Fig.~}

\newcommand{\rf}{F}
\newcommand{\rl}{L}
\newcommand{\om}{O}
\newcommand{\cm}{M}

\newcommand{\hc}{C}

\newcommand{\qd}{Q}

\newcommand{\coll}{W}
\newcommand{\pdf}{\mathbf{pdf}}

\newcommand{\argmax}[1]{\underset{#1}{\operatorname{argmax}}\medspace}

\graphicspath{{images/}}

\title{\LARGE \bf
Learning and Inference of Dexterous Grasps for Novel Objects with Underactuated Hands
}

\author{Marek Kopicki$^{1}$ and Carlos J. Rosales$^{2}$ and Hamal Marino$^{2}$ and Marco Gabiccini$^{2}$ and Jeremy L. Wyatt$^{1}$
\thanks{*This work was supported by EC-FP7-ICT-600918, PacMan.}
\thanks{$^{1}$Marek Kopicki and Jeremy L. Wyatt are with the Intelligent Robotics Laboratory, School of Computer Science,
        University of Birmingham, Birmingham, B15 2TT, UK.}
\thanks{$^{2}$Carlos J. Rosales, Hamal Marino and Marco Gabiccini are with Centro Piaggio, Universita di Pisa, Pisa, Italy.}
\thanks{Correspond to: {\tt\small M.S.Kopicki at cs.bham.ac.uk}}%
}
\begin{document}

\maketitle
\thispagestyle{empty}
\pagestyle{empty}

\begin{abstract}
Recent advances have been made in learning of grasps for fully actuated hands. A typical approach learns the target locations of finger links on the object. When a new object must be grasped, new finger locations are generated, and a collision free reach-to-grasp trajectory is planned. This assumes a collision free trajectory to the final grasp. This is not possible with underactuated hands, which cannot be guaranteed to avoid contact, and in fact exploit contacts with the object during grasping, so as to reach an equilibrium state in which the object is held securely. Unfortunately these contact interactions are i) not directly controllable, and ii) hard to monitor during a real grasp. We overcome these problems so as to permit learning of transferrable grasps for underactuated hands. We make two main technical innovations. First, we model contact interactions during the grasp implicitly. We do this by modelling motor commands that lead reliably to the equilibrium state, rather than modelling contact changes themselves. This alters our reach-to-grasp model. Second, we extend our contact model learning algorithm to work with multiple training examples for each grasp type. This requires the ability to learn which parts of the hand reliably interact with the object during a particular grasp. Our approach learns from a rigid body simulation. This enables us to learn how to approach the object and close the underactuated hand from a variety of poses. From nine training grasps on three objects the method transferred grasps to previously unseen, novel objects, that differ significantly from the training objects, with an 80\% success rate. 
\end{abstract}

\section{INTRODUCTION}
\label{sec:introduction}
\input{./inputTex/introduction.tex}


\section{OVERVIEW OF APPROACH}
\label{sec:overview}
\input{./inputTex/overview.tex}

\section{BASIC REPRESENTATIONS}
\label{sec:representations}
\input{./inputTex/representations.tex}

\section{DATA GENERATION}
\input{./inputTex/learning.tex}

\section{INFERENCE}
\label{sec:infer}
\input{./inputTex/inference.tex}

\section{RESULTS}
\label{sec:results}
\input{./inputTex/results.tex}


\input{./soft-grasp-arxiv2016.bbl}
\addtolength{\textheight}{-12cm}

%
%

\end{document}

%% file: inputTex/introduction.tex
Transferring dexterous grasps to novel objects is a challenging problem. One approach is to machine learn solutions with techniques able to perform powerful generalisation. Another is to use an underactuated hand to cope with shape variation. In this paper we combine the benefits of both approaches by learning grasps for underactuated hands. Underactuated hands exploit the contacts that occur during grasping to achieve a wide variety of final grasp configurations. The final grasp configuration depends not only on the final hand pose, but also on the object shape, and on the reach to grasp trajectory. An interesting challenge is to use machine learning to exploit these interactions. The key technical challenge in applying machine learning to grasping with underactuated hands is to learn the right motor actions so as to learn contact interactions that will lead to a reliable final grasp.

\begin{figure}
  \centering
  \begin{tabular}{ccc}
  \includegraphics[width=0.45\linewidth]{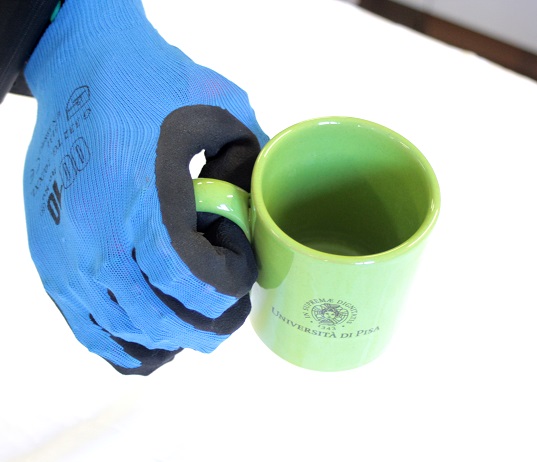} &
  \includegraphics[width=0.45\linewidth]{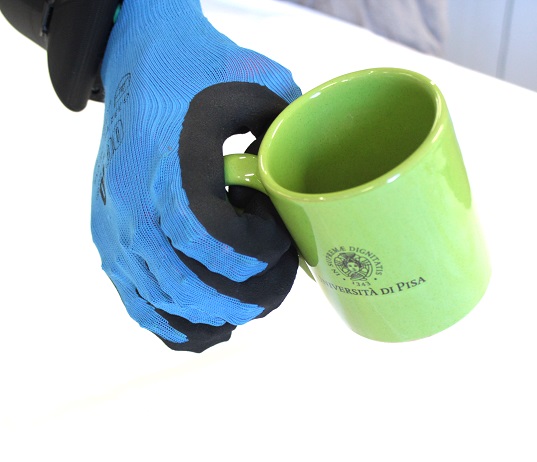} \\
  \end{tabular}
 \caption{{We want transferrable grasps that are robust to different initial hand-object poses, and thus different interactions during reach to grasp, thus reaching similar final grasp states. We achieve this by learning a set of trajectories that, associated with a model of the final grasp state, form an attractor basin around that state.}}
  \label{fig:two_grasps}
\end{figure}

One approach would be to ecplicitly learn the typical contact interactions that occur during a grasp, and to generate new grasps that reproduce these. The contact interactions are, however, complex and variable, even given small variations in object shape and friction. This makes transfer learning for underactuated hands challenging. There are two main novel technical contributions. First, we extend our learning method, so as to learn a grasp type from multiple trajectories. Second we implicitly encode the sequence of contact interactions by remembering the motor commands for finger closing as well as the wrist trajectory. We also demonstrate, for the first time, learning from examples generated in a rigid body physics simulation. Finally, at the grasp selection stage we now optimise across a space defined by several examples so as to maximise the chance of reaching a stable grasp. The method copes with partial and noisy shape information for the test objects. 


\subsection{Related Work}
Previous work in learning generalisable grasps falls broadly into two classes. One class of approaches utilises the shape of common object parts or their appearance to generalise grasps across object categories \cite{saxena2008b,detry2013a,herzog2014a, kroemer2012a}. This works well for low DoF hands. Another class of approaches captures the global properties of the hand shape either at the point of grasping, or during the approach \cite{ben2012generalization}. This global hand shape can additionally be associated with global object shape, allowing generalisation by warping grasps to match warps of global object shape \cite{hillenbrand2012transferring}. This second class works well for high DoF hands, but generalisation is more limited. We have previously achieved the advantages of both classes, generalising grasps across object categories with high DoF hands. In this paper we go beyond this, learning and generalising grasps for under-actuated hands.

Several hands with such behavior have been proposed in the literature with different implementations~\cite{Catalano2014Adaptive, Dollar2010Highly}, with a common goal: simplicity plus robustness. Their initial tests under human operation are promising, but autonomous grasping with underactuated hands faces challenges due to the almost non-observability of the finger deformation when the hand is constrained by the environment and/or a target object. Most of the existing planning algorithms for this type of hands boil down to generating good wrist poses and let the adaptive mechanism handle all variation and uncertainty while closing, such as~\cite{Eppner2015Planning}, where a sequence of wrist and object poses and the corresponding interaction wrenches are generated, which are expected to exploit environmental constraints. Another approach is that by~\cite{Bonilla2015Grasp}, where static wrist poses are sampled using different strategies around the object from where the fingers are closed using a rigid-body simulator, to finally select the areas of major success rate to generate new wrist poses.

While these approaches exploit, to some extent, the adaptive properties of the underactuated mechanism, they can be improved on. In this paper we show how we can, for the first time, learn grasps for underactuated hands that are then transferred to novel objects. This requires learning representations of the final grasp state that are amenable to transfer to new objects, grouping example grasps by the end grasp state, and learning and optimisation of reach-to-grasp trajectories.

%% file: inputTex/overview.tex
In our approach the main steps are as follows. A model of a training object is presented in a rigid body physics simulator. Then a number of example grasps are executed by a human, with the precise motions of hand and object during contact being determined by the simulation. Each example grasp continues until a final stable grasp state is reached. We call this the {\em equilibrium state}, consisting of the final hand shape, and the final set of contact relations between hand and object. For training and inference purposes each example grasp has two parts: an equilibrium state, and the reach to grasp trajectory leading to it.

We generate the example grasps in sets. Each set corresponds to a type of grasp, e.g. power or pinch. This means that the equilibrium states are similar within a set, but differ substantially between sets. 

Models are then learned for each grasp and for each set. Models are learned of the reach to grasp, the hand configuration in the equilibrium state, and the contact relations between hand and object in the equilibrium state. Given these models, when a new object is presented, a (partial) model of that object is obtained by sensing. This model is combined with the models learned from the training grasps.

Then many candidate equilibrium states, and associated candidate reach to grasp trajectories are generated by sampling. Finally they are optimised so as to maximise the likelihood of the grasp according to a product of experts.

%% file: inputTex/representations.tex
We now sketch the representations underpinning our approach. We define several models: an object model (partial and acquired from sensing); a model of the contact between a finger link and the object; a model of the whole hand configuration; and a model of the reach to grasp trajectory. First we describe the kernel density representation for all these models. Then we describe the surface features we use to encode some of these models. Then we follow with a description of each model type. Throughout we assume that the robot's hand comprises $N_L$ rigid \emph{links}: a palm, and several phalanges or links. We denote the set of links $L =\{L_i\}$. 

\subsection{Kernel Density Estimation}
\label{sec:kde}
$SO(3)$ denotes the group of rotations in three dimensions. A feature belongs to the space $SE(3) \times \mathbb R^2$, where $SE(3) = \mathbb R^3 \times SO(3)$ is the group of 3D \emph{poses}, and surface descriptors are composed of two real numbers. We extensively use probability density functions (PDFs) defined on $SE(3) \times \mathbb R^2$.  We represent these PDFs non-parametrically with a set of $K$ features (or particles) $x_j$
\begin{equation}
S = \left\lbrace x_j : x_j \in \mathbb R^3 \times SO(3) \times \mathbb R^2 \right\rbrace_{j \in [1,K]}.
\end{equation}
The probability density in a region is determined by the local density of the particles in that region. The underlying PDF is created through \emph{kernel density estimation} \cite{silverman1986a}, by assigning a kernel function $\mathcal{K}$ to each particle supporting the density, as
\begin{equation}\label{eq:d}
\pdf(x) \simeq \sum_{j=1}^K w_j \mathcal{K}(x| x_{j}, \sigma),
\end{equation}
where  $\sigma \in \mathbb R^3$ is the kernel bandwidth and $w_j \in \mathbb R^{+}$ is a weight associated to $x_j$ such that $\sum_j w_j = 1$. We use a kernel that factorises into three functions defined by the separation of $x$ into $p \in \mathbb R^3$ for position, a quaternion $q \in SO(3)$ for orientation, and $r \in \mathbb R^2$ for the surface descriptor. Furthermore, let us denote by $\mu$ another feature, and its separation into position, orientation and a surface descriptor. Finally, we denote by $\sigma$ a triplet of real numbers:
\begin{subequations}
\begin{align}
x &= (p, q, r),\\
\mu &= (\mu_p, \mu_q, \mu_r),\\
\sigma &= (\sigma_p, \sigma_q, \sigma_r).
\end{align}
\label{eq:feature}
\end{subequations}
We define our kernel as
\begin{equation}\label{eq:kernel_in_se3}
\mathcal{K}(x| \mu, \sigma) = \mathcal{N}_3(p| \mu_p, \sigma_p) \Theta(q| \mu_q, \sigma_q) \mathcal{N}_2(r| \mu_r, \sigma_r)
\end{equation}
where $\mu$ is the kernel mean point, $\sigma$ is the kernel bandwidth, $\mathcal{N}_n$ is an $n$-variate isotropic Gaussian kernel, and ${\Theta}$ corresponds to a pair of antipodal von Mises-Fisher distributions which form a Gaussian-like distribution on $SO(3)$ \cite{fisher1953a,sudderth2006a}. The value of ${\Theta}$ is given by
\begin{equation}
\Theta(q|\mu_q, \sigma_q) = C_4(\sigma_q) \frac {e^{\sigma_q \; \mu_q^T q} + e^{-\sigma_q \; \mu_q^T q}}2
\end{equation}
where $C_4(\sigma_q)$ is a normalising constant, and $\mu_q^T q$ denotes the quaternion dot product.

Using this representation, conditional and marginal probabilities can easily be computed from \eq\eqref{eq:d}. The marginal density $\pdf(r)$ is computed as
\begin{multline}\label{eq:marginal}
\pdf(r)\\
= \iint \sum_{j=1}^K w_j \mathcal{N}_3(p| p_i, \sigma_p) \Theta(q| q_i, \sigma_q) \mathcal{N}_2(r| r_i, \sigma_r) \textnormal{d}p\textnormal{d}q \\
=  \sum_{j=1}^K w_j \mathcal{N}_2(r| r_j, \sigma_r),
\end{multline}
where $x_j = (p_j, q_j, r_j)$.
The  conditional density $\pdf(p,q|r)$ is given by
\begin{multline}\label{eq:conditional}
\pdf(p,q|{r}) = \frac{\pdf(p, q, {r})}{\pdf({r})}\\
= \frac{\sum_{j=1}^K w_j \mathcal{N}_2({r}| r_j, \sigma_r) \mathcal{N}_3(p| p_j, \sigma_p) \Theta(q|q_j, \sigma_q)}{\sum_{j=1}^K w_j \mathcal{N}_2({r}| r_j, \sigma_r)}. 
\end{multline}

\subsection{Surface Features}
\label{sec:surface_features}

All objects considered in the paper are represented by point clouds for the purpose of learning and testing. Test object models were constructed from a single view with a depth camera, and were thus incomplete5. We directly augment these points with a frame of reference and a surface feature that is a local curvature descriptor. For compactness, we also denote the pose of a feature as $v$. As a result,
\begin{equation}
x = (v, r), \qquad v = (p, q).
\label{eq:surface.feature}
\end{equation}

The surface normal at $p$ is computed from the nearest neighbours of $p$ using a PCA-based method (e.g. \cite{kanatani2005statistical}). The surface descriptors are the local \emph{principal curvatures} \cite{spivak1979comprehensive}. Their directions are denoted $k_1, k_2 \in \mathbb R^3$, and the curvatures along $k_1$ and $k_2$ form a $2$-dimensional feature vector $r = (r_1, r_2) \in \mathbb R^2$. 
The surface normal and the principal directions define the orientation $q$ of a frame that is associated with the point $p$. 

\subsection{Object Model}
\label{sec:object_model}

Thus, given a point cloud of object $n = 1..N_O$, a set of $K_{O_n}$ features $\lbrace (v_{jn}, r_{jn}) \rbrace$ can be computed. This set of features defines, in turn, a joint probability distribution, which we call the \emph{object model}:
\begin{equation}
\om_n(v, r) \equiv \pdf^\om(v, r) \simeq \sum_{j=1}^{K_{O_n}} w_{jn} \mathcal{K}(v, r|{x_{jn}}, \sigma_{x})
\label{eq:om}
\end{equation}
where $\om$ is short for $\pdf^\om$, $x_{jn} = (v_{jn}, r_{jn})$,  $\mathcal{K}$ is defined in \eq\eqref{eq:kernel_in_se3} with bandwidth $\sigma_{x} = (\sigma_{v}, \sigma_{r})$, and where all weights are equal, $w_{jn} = 1/{K_{O_n}}$. In summary this object model $\om$ represents the object as a pdf over surface normals and curvatures.

\section{LEARNED MODELS}

We now describe the representations for each of the three models that are learned from a set of grasp examples. We start with the contact receptive field, proceed with the contact model, the equilibrium state hand configuration model, and finish with the reach to grasp model.

\subsection{Contact Receptive Field}\label{sec:contact_recfield}

\begin{figure}[t]
\centerline{\includegraphics[width=6cm]{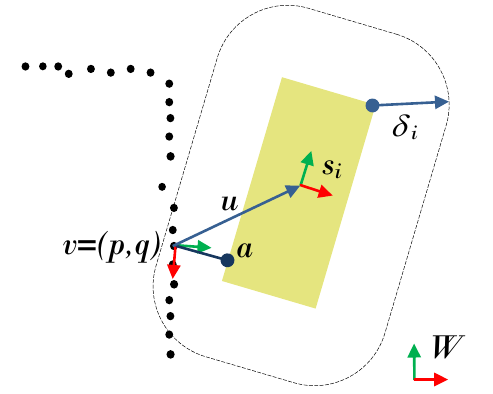}}
\caption[Contact receptive field]{The contact receptive field $\rf$ associated with the $i$-th link $\rl_i$ (solid yellow block) with link pose $s_i$. The black dots are samples from the surface of an object. The distance $a$ between feature $v$ and the closest point $a$ on the link's surface is shown. The rounded rectangle illustrates the cut-off distance $\delta_i$. The poses $v$ and $s_i$ are expressed in the world frame $W$. The arrow $u$ is the pose of $\rl_i$ relative to the frame for the surface feature $v$.}
\label{fig:contact_recfield}
\end{figure}

The \textit{contact receptive field} $\rf_i$ is a region of space relative to the associated robot link $\rl_i$ (see \fig\ref{fig:contact_recfield}) which specifies the neighbourhood of a particular robot link. The contact receptive field $\rf_i$ is realised as a function of surface feature pose $v$:
\begin{equation}
\rf_{i} : SE(3) \rightarrow [0, 1]
\label{eq:contact_recfield_model}
\end{equation}
the value of which determines the relevance of a particular surface feature $x = (v, r)$ to a given link $\rl_i$ in terms of the likelihood of the physical contact. We use contact receptive fields which are family of parameterised functions for which the value falls off quickly with the distance to the link:
\begin{equation}
\rf_{i}(v|\lambda_i,\delta_i) = \begin{cases}\exp(-\lambda_i ||p-a||^2) \quad &\textnormal{ if } ||p-a|| < \delta_i\\
0 \quad &\textnormal{ otherwise},\end{cases}
\label{eq:contact_recfield_func}
\end{equation}
where $\lambda_i \in \mathbb R^{+}$ and $a$ is the point on the surface of $\rl_i$ that is closest to $v = (p, q)$. This means that the contact contact receptive field will only take account of the local shape, while falling off smoothly. A variety of monotonic, fast declining functions could be used instead.

\subsection{Contact Model}\label{sec:contact.model}

Let us assume that we have as many grasp examples of the same grasp type $g$ as the number of objects $N_O$. We denote by $u_{ij} = (p_{ij}, q_{ij})$ the pose of $\rl_i$ relative to the pose $v_j$ of the $j$-th object feature. In other words, $u_{ij}$ is defined as
\begin{equation}
u_{ij} = v_j^{-1} \circ s_i,
\label{eq:local.pose}
\end{equation}
where $s_i$ denotes the pose of $\rl_i$, $\circ$ denotes the pose composition operator, and $v_j^{-1}$ is the inverse of $v_j$, with $v_j^{-1} = (-q_j^{-1}p_j, q_j^{-1})$ (see \fig\ref{fig:contact_recfield}). 

Contact model $\cm_{in}$ encodes the joint probability distribution of surface features of object $n$ and of the 3D pose of the $i$-th hand link in the equilibrium state. Let us consider the hand as having grasped given training object $n$. The contact model for link $\rl_i$ is denoted by
\begin{equation}\label{eq:M}
\cm_{in}(U, R) \equiv \pdf^\cm_{in}(U, R)
\end{equation}
where $\cm_i$ is short for $\pdf^\cm_{in}$, $R$ is the random variable modelling surface features of object $n$, and $U$ models the pose of $\rl_i$ \emph{relative} to the frame of reference defined by the directions of principal curvature and the surface normal. In other words, denoting realisations of $R$ and $U$ by $r$ and $u$, $\cm_{in}(u, r)$ is proportional to the probability of finding $\rl_i$ at pose $u$ relative to the frame of a nearby object surface patch that exhibits features (principal curvatures) equal to $r$.

The contact model of object $n$ is estimated as
\begin{equation}\label{eq:cm_n}
\begin{split}
&\cm_{in}(u, r) \simeq \frac 1Z \int \mathcal{K}(u|v^{-1}s_i, \sigma_\cm) \om_{n}(v, r) \rf_{i}(v) dv\\
&= \frac 1Z \sum_{j=1}^{K_{O_n}} w_{jn} \mathcal{K}(u|v_{jn}^{-1}s_i, \sigma_v) \mathcal{N}_2({r}| r_{jn}, \sigma_r) \rf_{i}(v_{jn})
\end{split}
\end{equation}
where $Z \in \mathbb R^+$ is a normalising constant and $\mathcal{K}$ is kernel function \eqref{eq:kernel_in_se3} defined at poses from \eq\eqref{eq:local.pose}, and where we have performed integration over all kernels of the object model \eqref{eq:om} which uniquely determine poses $v$.

We also introduce the idead of a \textit{contact model norm}, which estimates the likelihood of a physical contact of link $i$ with surface features of object $n$:
\begin{equation}\label{eq:cm_norm}
\|\cm_{in}\| = \frac 1Z \sum_{j=1}^{K_{O_n}} w_{jn} \rf_{i}(v_{jn})
\end{equation}
where $Z \in \mathbb R^+$ is a normalising constant. We use this norm to help estimate which links are reliably involved in an grasp equilibrium state when there are several example grasp trajectories.

\subsection{Contact Model Selection}

The overall number of non-empty contact models $\cm_{in}$ for some example grasp $n$ is usually smaller than the number of links $N_L$ of the robot hand. The contact selection procedure determines which contact models should be instantiated for a given grasp type. The issue is complex because we require multiple grasp trajectories to learn from. Because of the variability in how the hand and object interact, it is then non-trivial to determine which finger links reliably participate in the equilibrium state of the grasp. The contact selection procedure determines this. The procedure starts with comparing contact model norms \eqref{eq:cm_norm} to the average model norm for all grasp examples and all robot links. The results of this comparison are stored in binary variables $b_{in}$:
\begin{equation}
b_{in} = \begin{cases} 1 \quad &\textnormal{ if } \frac{N_L N_O \|\cm_{in}\|}{\sum_{jk}\|\cm_{jk}\|} > \eta_i \\
0 \quad &\textnormal{ otherwise},\end{cases}
\label{eq:cm_n_binary}
\end{equation}
where $\eta_i \in \mathbb{R}^+$ is a threshold constant, $N_L$ is the number of hand links and $N_O$ is the number of training objects.

The contact model $\cm_i$ is instantiated if the total number of non-empty contact models, determined by $b_{in}$, is higher than some minimum value given the total number of objects or training grasps $N_O$. This is encoded in binary variable $c_i$:
\begin{equation}
c_i = \begin{cases} 1 \quad &\textnormal{ if } \frac{1}{N_O} \sum_k b_{ik} > \zeta\\
0 \quad &\textnormal{ otherwise},\end{cases}
\label{eq:cm_binary}
\end{equation}
where $\zeta \in \mathbb{R}^+$ is a threshold constant. The contact model $\cm_i$ is then constructed as a mixture of $\cm_{in}$ as follows:
\begin{equation}\label{eq:cm}
\cm_{i}(u, r) = \sum_{n=1}^{N_O} c_{i} b_{in} \cm_{in}(u, r)
\end{equation}

The above procedure is performed independently for each grasp type $g$. Each grasp type may also involve a different set of objects. We denote the set of contact models learned for a particular grasp type $g$ as $\mathcal{\cm}^g=\{\cm^g_i\}$. The parameters $\lambda$, $\eta$, $\zeta$ and $\sigma_{p}$, $\sigma_{q}$, $\sigma_{r}$ were chosen empirically. The time complexity for learning each contact model $\cm_{in}(u, r)$ from an example grasp is $\Omega(T_i K_{O_n})$ where $T_i$ is the number of triangles in the tri-mesh describing hand link $i$, and $K_{O_n}$ is the number of points of object $n$.

\subsection{Equilibrium State Hand Configuration Model}

The equilibrium state hand configuration model, denoted $h^e_c(j) \in \mathbb R^D$, for the grasp examples $n = 1 \ldots N_O$ within the set of $n$ training examples of a particular grasp type $g$. The purpose of this model is to restrict the grasp search space (during grasp transfer) to hand configurations that resemble those observed during training. We combine the configurations for the examples $n = 1 \ldots N_O$ to create a single mixture model density:
\begin{equation}
\hc^g(h^e_c) \equiv \sum_{n=1}^{N_O} \mathcal{N}_D(h^e_c|h^e_c(n), \sigma_{h^e_c}) 
\label{eq:hc}
\end{equation}
This expresses a density over hand configurations in the equilibrium state for a grasp type $g$.

\subsection{Reach to Grasp Model}

For a particular grasp type, in addition to modelling the equilibrium states of the hand, we must also model the trajectories taken to reach those equilibrium states. A single reach to grasp trajectory for an underactuated hand has three elements: the tool centre point (wrist) trajectory, the hand configuration trajectory, and the motor signal trajectory. We assume that a trajectory starts at time $t_0$ and ends in the equilibrium state at time $t_e$. We denote the wrist trajectory $h_w^{0:e}$, the hand configuration trajectory $h_c^{0:e}$, and the motor signal trajectory $h_m^{0:e}$ respectively. The motor signal can be a wide variety of signals in practice. Here we choose it to be the position of the single actuator. When the hand is not in contact with an object the motor signal and the wrist pose together determine the hand configuration. When in contact, the actual hand configuration will differ. The reach to grasp model is simply the concatenation of each component $(h_w^{0:e} , h_c^{0:e} , h_m^{0:e})$. The set of reach to grasp trajectory models define an attractor basin, leading towards the final hand configuration. Although this notion of using the motor command sequence is simple, it is essential to encode the envelope of hand behaviours that will lead to a stable equilibrium grasp.

In the next section we explain how we gather the grasp examples that are used to learn these models. Then in Section~\ref{sec:infer} the inference method---by which the models are used to generate grasps for new objects---is described.

%% file: inputTex/learning.tex

There are several ways to implement underactuation in a dexterous hand. In this paper we employ an approach based on adaptive synergy transmission, due to its simplicity and robust design, and its ability for complex interaction with the environment. The Pisa/IIT SoftHand~\cite{Catalano2014Adaptive} implements such a transmission mechanism. This hand has 19 degrees of freedom (DoF) distributed over four fingers and an opposable thumb, but only 1 degree of actuation (DoA). The synergy motion of the hand in free space has been derived from databases of human hand postures. The overall behaviour parameters are the matrices that correspond to the transmission ratio,~$R$, to the joint stiffness,~$K_q$. The actuation is done through a single tendon routed throuh all joints, making the fingers flex and abduct.


Moving such a hand to grasp an object results in a hard-to-predict contact and hand shapes due to the adaptivity. We thus generate a variety of grasp examples to cover a portion of the interaction space. However, recording many trajectories of all the finger elements that affect the grasp in the real world is non-trivial. For this reason, we generate the example interactions for training using a rigid-body physics simulator, where these problems are avoided. The main two simulation elements we have developed are the contact stability model and the hand behavior model. In the case of the Pisa/IIT softhand, the latter depends heavily on the former. We used the standard distribution of Gazebo and Open Dynamic Engine, both in widespread use. The adaptive synergy equations have been implemented as a plugin to these, and accompany the proper kinematic description of the Pisa/IIT SoftHand\footnote{The Pisa/IIT SoftHand ROS/Gazebo packages are available at \url{https://github.com/CentroEPiaggio/pisa-iit-soft-hand}}.

\begin{figure}
  \centering
  \begin{tabular}{ccc}

  \includegraphics[width=0.3\linewidth]{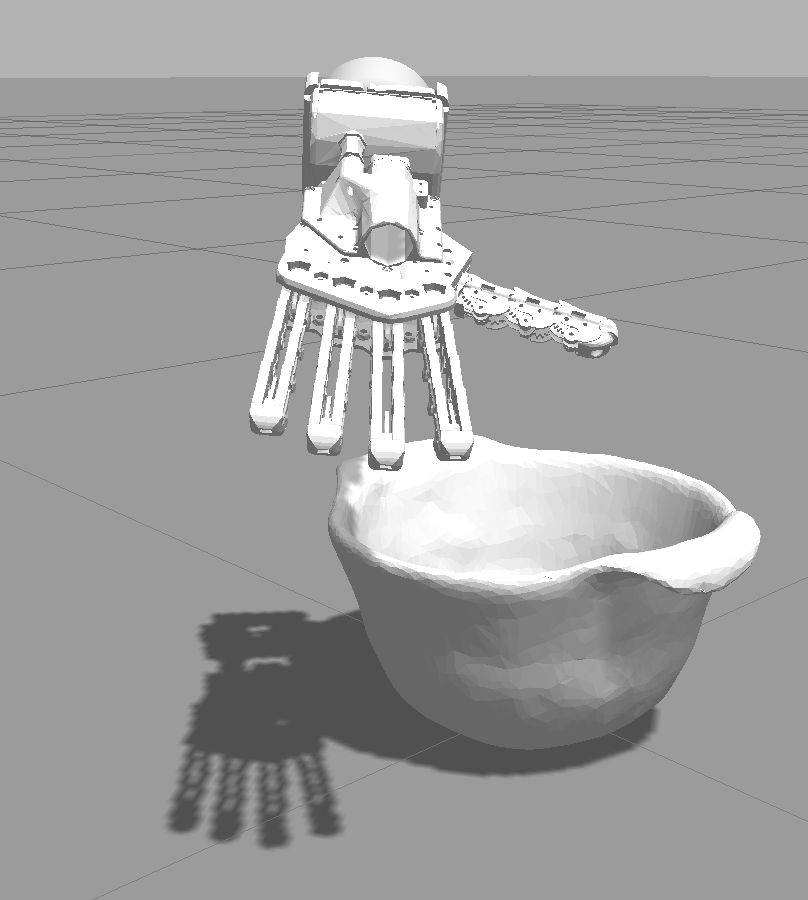} &

  \includegraphics[width=0.3\linewidth]{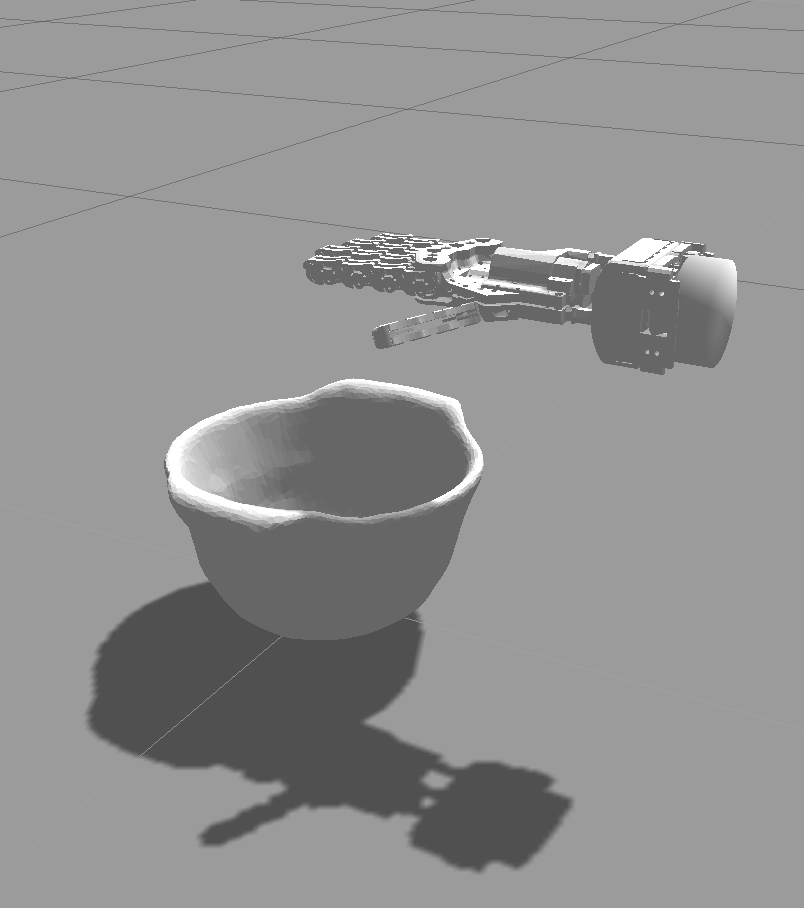} &

  \includegraphics[width=0.3\linewidth]{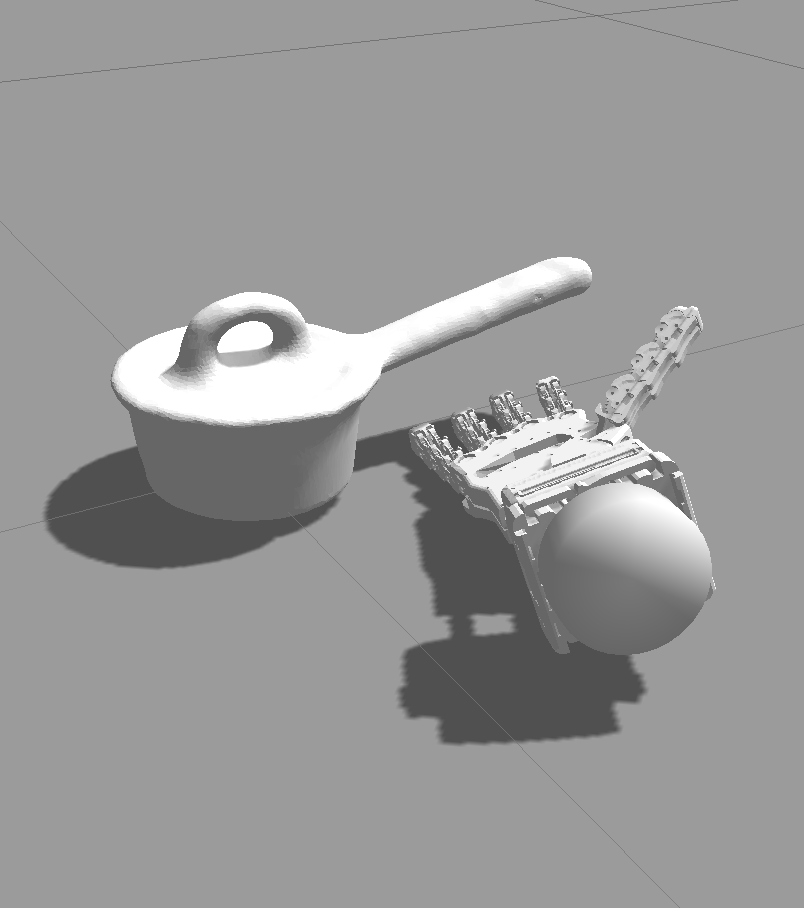} \\

  \includegraphics[width=0.3\linewidth]{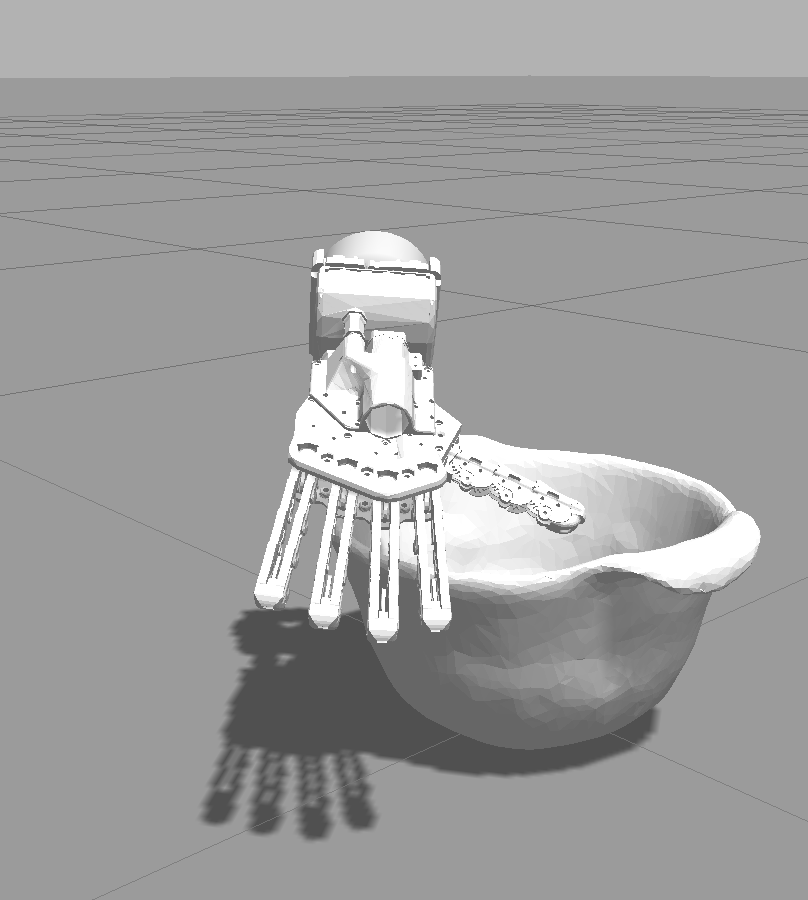} &

  \includegraphics[width=0.3\linewidth]{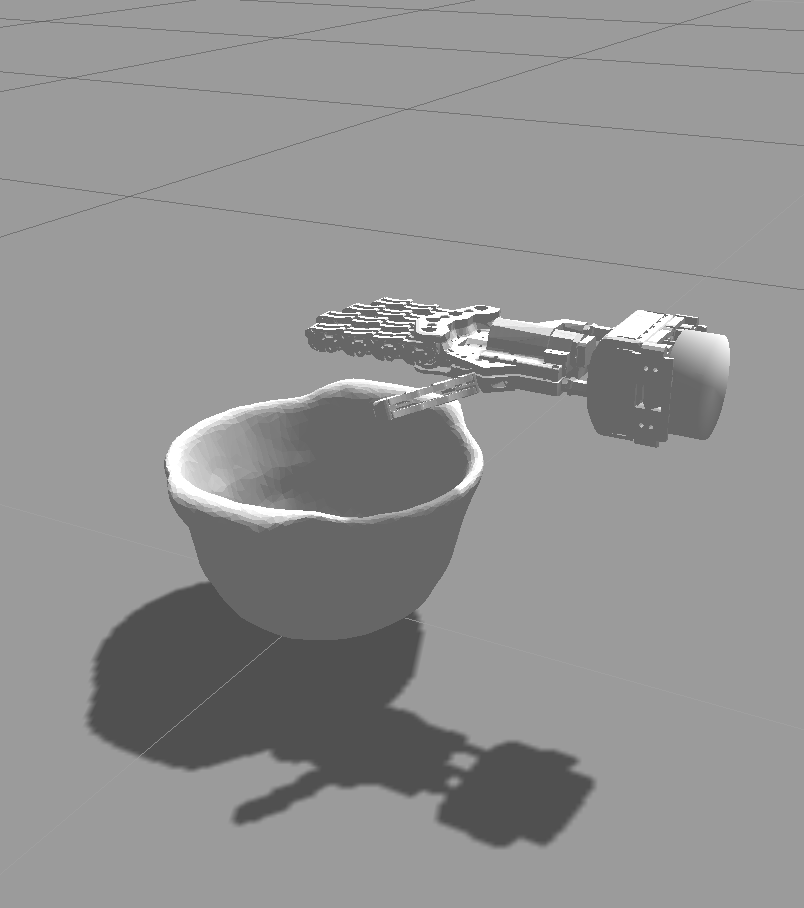} &

  \includegraphics[width=0.3\linewidth]{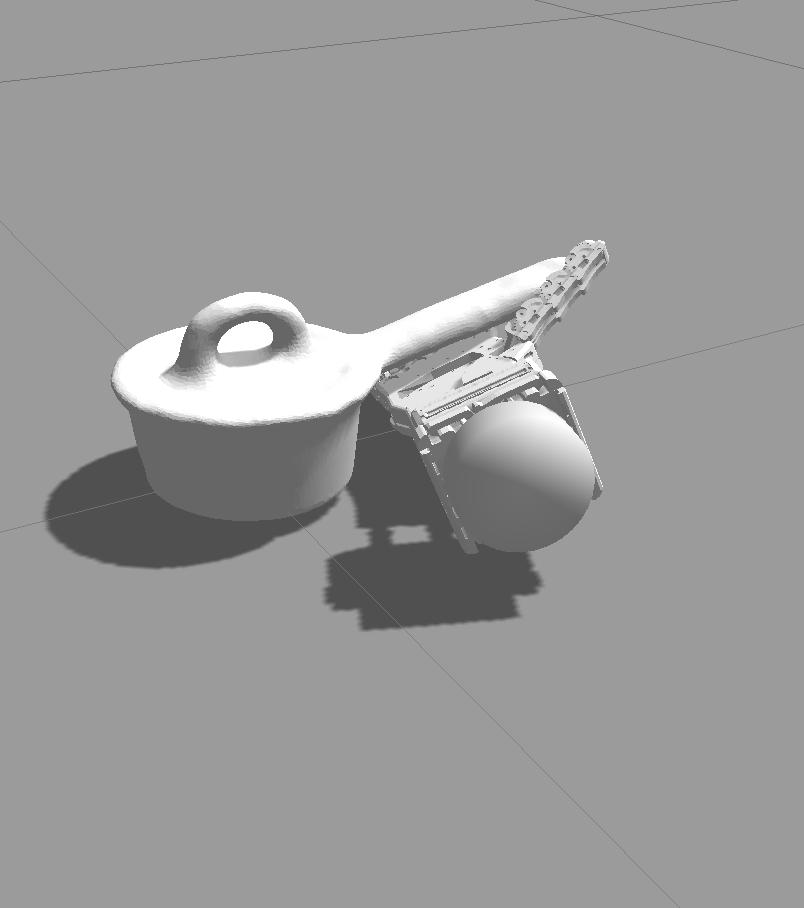} \\

  \includegraphics[width=0.3\linewidth]{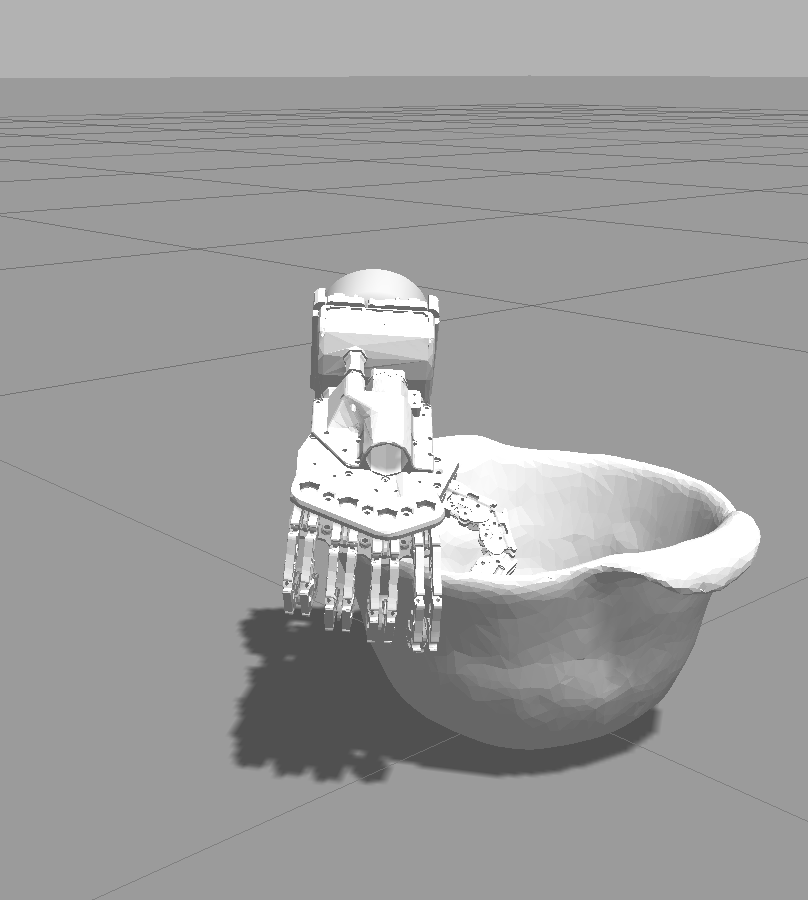} &

  \includegraphics[width=0.3\linewidth]{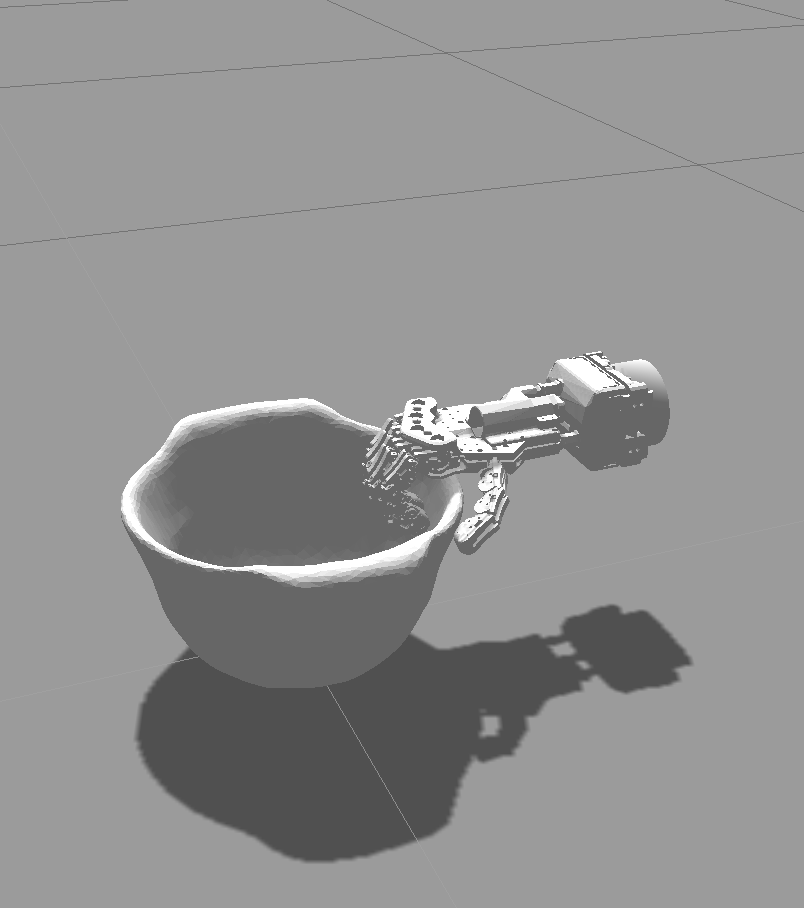} &

  \includegraphics[width=0.3\linewidth]{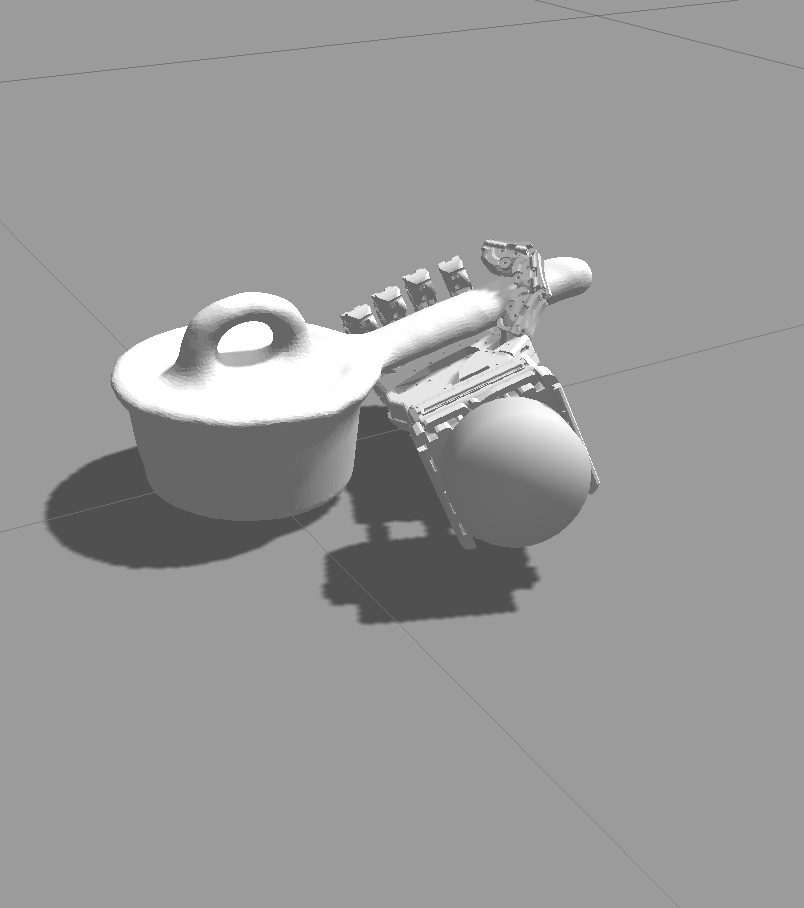} \\

  \end{tabular}

  \caption{Snapshots of the simulation of a pinch and rim grasp types for the colander (first and second columns), and handle grasp for the pot (last column).}
  \label{fig:simulations}
\end{figure}

%
At the current state, there are no generally accepted measures concerning whether a grasp by an underactuated hand  is good or not, hence the lack of robust grasp planners for them is not a surprise. Thus, generating a large dataset at this point is useless, and there are plans in the future to cover this area. As a result, we generated the examples by guiding manually the hand to a ``nice'' grasp as shown in Fig.~\ref{fig:simulations}. In this simpler scenario, we assume without loss of generality that the grasps are labelled by type. In our example dataset, we have three grasp types namely pinch, rim and by-the-handle. The main difference between pinch and rim is the fingers configuration w.r.t. the top border of objects. In the pinch grasp, the thumb goes inside whereas in the rim grasp, the fingers go inside. In the latter, the container can be filled with liquid while holding, for instance.
For each grasp example, the corresponding dataset comprises the set of trajectories as described in the previous section.





%% file: inputTex/inference.tex
After acquiring the models from a set of training grasps, we present the robot with a test (query) object. The goal is to find a generalisation of the training grasp that is likely according to all of the model types simultaneously. First of all, we obtain a point cloud for the test object, and thus an object model. We then combine every contact model with that object model, so as to obtain a set of {\em query densities}, one for each link with a contact model defined for the example grasp. The $i$-th query density $\qd_i$ is a density modelling where the $i$-th link can be placed, in the equilibrium state, with respect to the surface of a new object. 

From the query densities, a candidate equilibrium grasp state is generated. This is then augmented with a reach to grasp trajectory that finishes close to the candidate equilibrium grasp state. Finally we refine the equilibrium grasp and reach to grasp by performing a simulated annealing search in the space of equilibrium state wrist poses and hand configurations, so as to maximise the grasp likelihood. We repeat the entire process many times. This procedure generates many possible grasps, ranked by likelihood. We give details below.

\subsection{Query Density}

A query density is, for a hand link and an object model, a density over the pose of that hand link relative to the object. Intuitively the query density encourages a finger link to make contact with the object at locations that have similar local surface curvature to that in the training example. Specifically, we use $K_{Q_i}$ kernels centred on the set of weighted finger link poses:
\begin{equation}
\qd_i(s) \simeq \sum^{K_{Q_i}}_{j=1} w_{ij} \mathcal{N}_3(p|{\hat{p}_{ij}}, \sigma_{p}) \Theta(q|{\hat{q}_{ij}}, \sigma_{q})
\label{eq:qd.approx}
\end{equation}
with $j$-th kernel centre $({\hat{p}_{ij}}, {\hat{q}_{ij}}) = \hat{s}_{ij}$, and weights are normalised $\sum_j w_{ij} = 1$. When a test object is presented, a set of query densities $\mathcal{Q}^g$ is calculated for the equilibrium state for each training grasp $b$ for the grasp type $g$. The set $\mathcal{Q}^g_b =\{\qd_{b,i}^g\}$ has $N^g_Q=N^g_M$ members, one for each contact model $M_i^g$ in $\mathcal{M}^g$. We estimate the query density using a Monte Carlo procedure detailed in Alg.\ref{alg:mc}.

\begin{algorithmbox}[floatplacement=t,float,fontupper=\small,label={alg:mc}]{Pose sampling ($\cm_i${,} $\om$)}
\begin{minipage}{\linewidth}\begin{tabbing}%
For \= samples $j=1$ to $K_{Q_i}$\\
  \> Sample $(\hat{v}_j, \hat{r}_j) \sim \om(v, r)$\\
  \> Sample from conditional density $(\hat{u}_{ij}) \sim \cm_i(u|\hat{r}_j)$\\
	\> Compute sample weight $w_{ij} = \cm_i(\hat{r}_j)$\\
	\> $\hat{s}_{ij} = \hat{v}_{j} \circ \hat{u}_{ij}$ \\
	\> separate $\hat{s}_{ij}$ into position $\hat{p}_{ij}$ and quaternion $\hat{q}_{ij}$\\
return $\{  (\hat{p}_{ij}, \hat{q}_{ij}, w_{ij} ) \}, \,  \forall j $
\end{tabbing}\end{minipage}
\end{algorithmbox}

\subsection{Equilibrium Grasp Generation} 
Once query densities have been created for the new object for each training example, an initial set of equilibrium state grasps is generated for each grasp type $g$. For each candidate equilibrium grasp of a particular grasp type we proceed as follows. First an example grasp is selected at random. Then a finger link is selected at random. This ‘seed’ link indexes its query density $\qd_i^g$. A link pose $s_i$ is then sampled from that query density. Then an equilibrium state hand configuration $h^e_c$ is sampled from $\hc^g$. Together the seed link and the hand configuration define a complete equilibrium state hand pose $h$ in the workspace via forward kinematics. This is an initial `seed' grasp, which will subsequently be refined. A large set of such initial solutions is generated, where $h_e^g(j)=(h_w^{e}(j) , h_c^{e}(j))$ means the $j^{th}$ initial solution for  grasp type $g$.

\subsection{Reach to Grasp Generation}
\begin{figure}

 \includegraphics[width=0.45\linewidth]{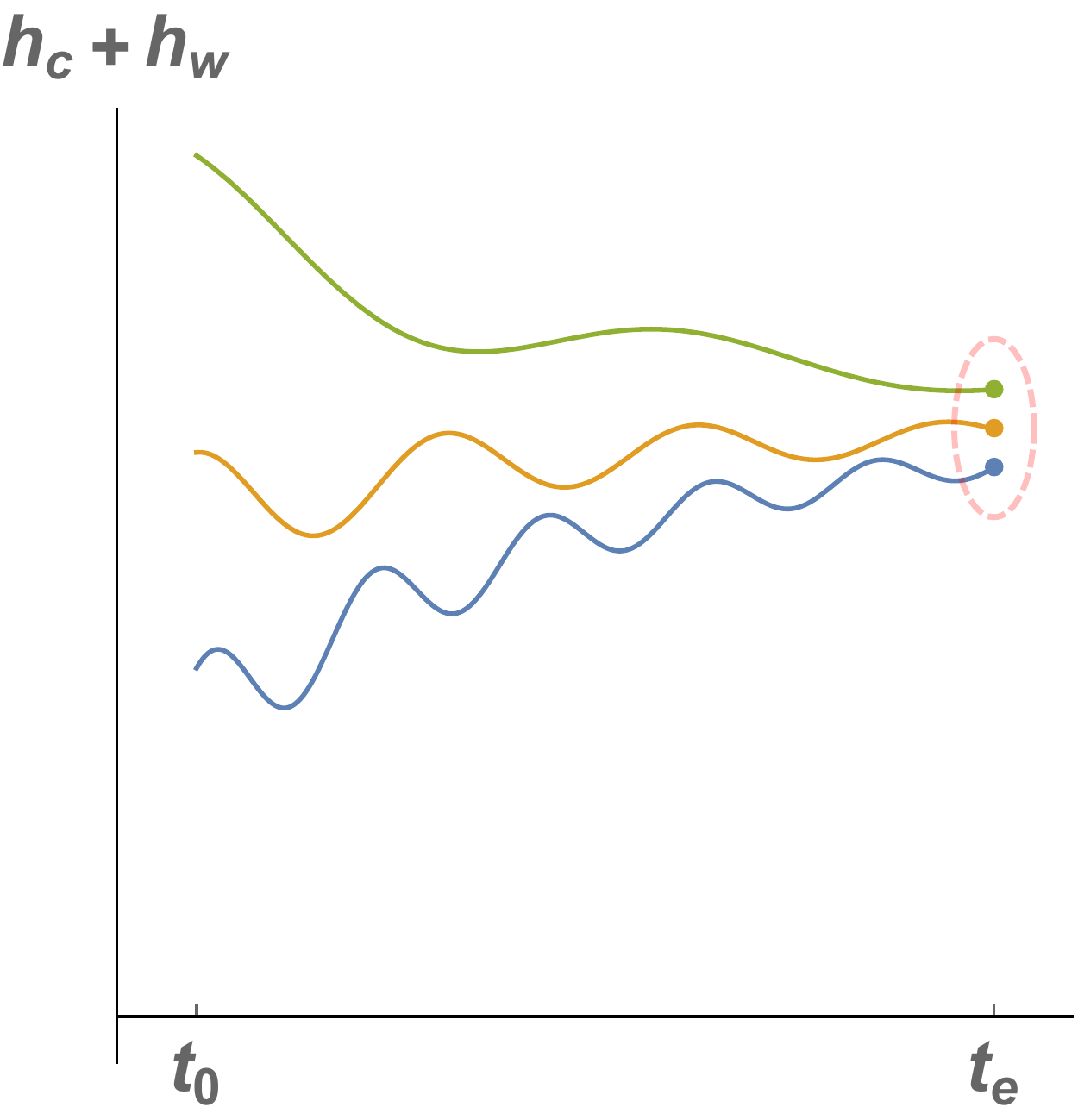}
  \includegraphics[width=0.45\linewidth]{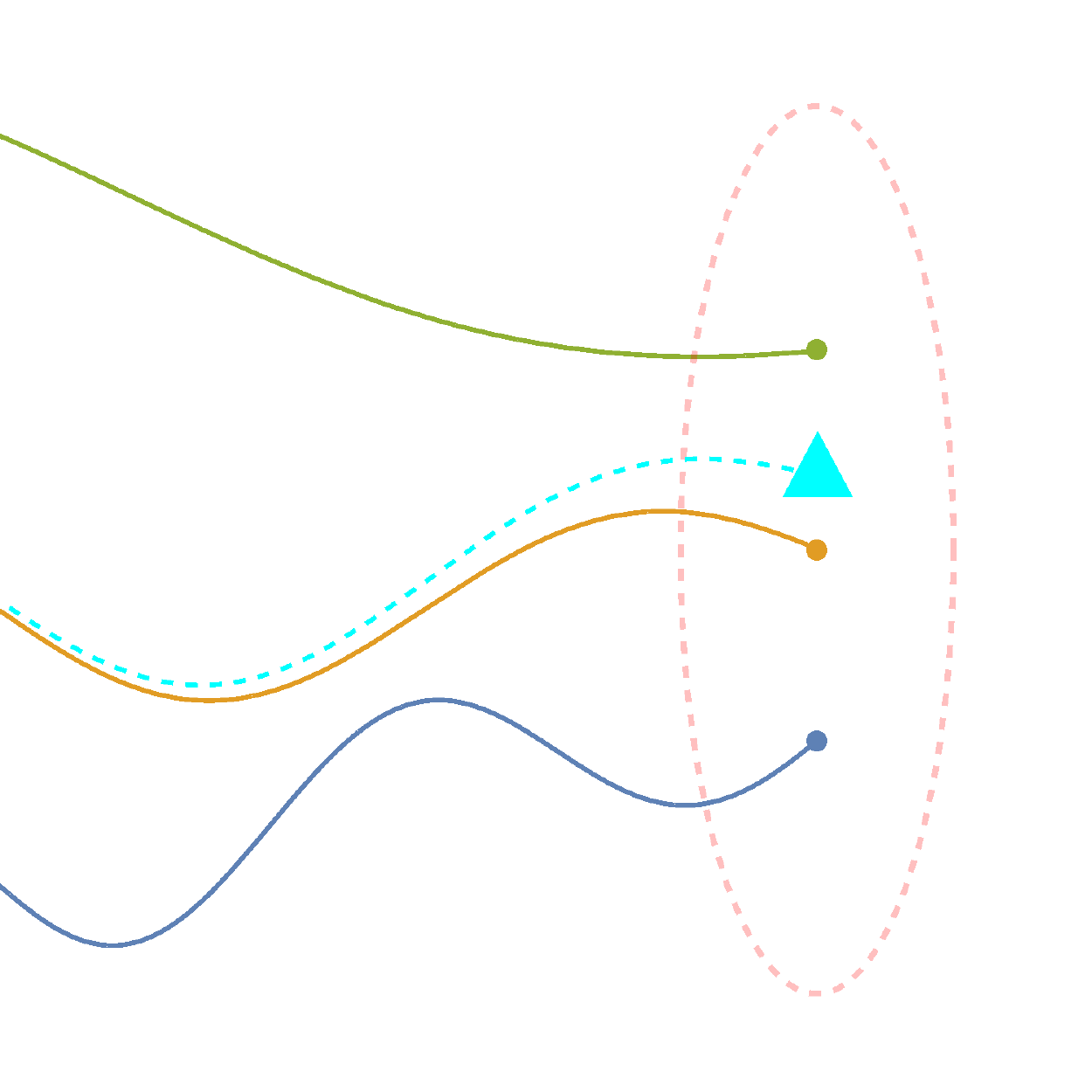}
 \caption{{(Left) We acquire set of reach to grasp trajectories that, associated with a model of the final equilibrium grasp state, form an attractor basin around that state. (Right) When an equilibrium state hand configuration is generated (triangle) a reach to grasp trajectory is sampled, and then the configuration component of the trajectory is smoothly interpolated between the selected reach to grasp and the generated equilibrium state grasp.}}
  \label{fig:reaches}
\end{figure}

Given an equilibrium grasp, a reach to grasp trajectory is selected and adapted to maximise the chance of reaching that equilibrium grasp state. Specifically, we sample a reach to grasp model $(h_w^{0:e} , h_c^{0:e} , h_m^{0:e})$ according to a multinomial probability distribution created by the normalised values of a Gaussian centred on the candidate equilibrium grasp $h_e^g(j)$. To align the selected reach to grasp to the candidate equilibrium grasp, the wrist trajectory $h_w^{0:e}$ is trivially redefined to be relative to the frame $h_w^{e}(j)$. Then the configuration trajectory only $h_c^{0:e}$ is warped (see Fig. \ref{fig:reaches}) so that it smoothly shifts from $h_c^{0}$ to $h_c^{e}(j)$ from the beginning to the end of the trajectory. 
Having generated an initial solution set $\mathcal{H}^{1}$ stages of optimisation and selection are then interleaved.

\subsection{Grasp Optimisation}
The objective of the grasp optimisation steps is, given a candidate equilibrium grasp and a reach to grasp model, to find a grasp that maximises the product of the likelihoods of the query densities and the hand configuration density
\begin{multline}
\argmax{(h)}  \mathcal{L}^g(h) = \argmax{(h)}  \mathcal{L}^g_\hc(h) \mathcal{L}^g_\qd(h) \\
 = \argmax{(h_w, h_c)}   \hc^{g}(h_c) \prod_{\qd_i^g \in \mathcal{Q}^g} \qd_i^g\left(k_{i}^{\mathrm{for}}\left(h_w, h_c\right)\right)
\label{eq:grasping.product}
\end{multline}
where ${\cal L}^g(h)$ is the overall likelihood, where $\hc^g(h_c)$ is the hand configuration model~\eqref{eq:hc}, $\qd_i^g$ are query densities~\eqref{eq:qd.approx}. Thus whereas each initial grasp is generated using only a single query density, grasp optimisation requires evaluation of the grasp against all query densities. It is only in this improvement phase that all query densities must be used. Improvement is by simulated annealing (SA) \cite{kirkpatrick83optimizationby}. The SA temperature $T$ is declined linearly from $T_{1}$ to $T_{K}$ over the $K$ steps. In each time step, one step of simulated annealing is applied to every grasp $m$ in $\mathcal{H}^k$.
 
 \subsection{Grasp Selection}
At predetermined selection steps (here steps 1 and 50 of annealing), grasps are ranked and only the most likely $10\%$ retained for further optimisation. During these selection steps the criterion in \eqref{eq:grasping.product} is augmented with an additional expert $\coll(h_w, h_c)$ penalising collisions for the entire reach to grasp trajectory in a soft manner. This soft collision expert has a cost that rises exponentially with the greatest degree of penetration through the object point cloud by any of the hand links. We thus refine \eq\ref{eq:grasping.product}:

\begin{multline}\label{eq:grasping.likelihood}
{\cal L}^g(h) = {\cal L}^g_\coll(h) {\cal L}^g_\hc(h) {\cal L}^g_\qd(h) = \\
= \coll(h_w, h_c) \hc^g(h_c) \prod_{\qd_i \in \mathcal{Q}^{g}} \qd^g_i\left(k_{i}^{\mathrm{for}}\left(h_w, h_c\right)\right)
\end{multline}

where ${\cal L}^g(h)$ is now factorised into three parts, which evaluate the collision, hand configuration and query density experts, all at a given hand pose $h$. A final refinement  of the selection criterion is due to the fact the number of links involved in a grasp varies across grasp types. Thus the number of query densities $N^{g_1}_\qd$, $N^{g_2}_\qd$ for different grasp models $g_1 \neq g_2$ also varies, and so the values of ${\cal L}^{g_1}$ and ${\cal L}^{g_2}$ cannot be compared directly. Given the grasp with the maximum number of involved links $N^{\max}_\qd$, we therefore normalise the likelihood value~\eqref{eq:grasping.likelihood} with
\begin{equation}
\left\|{\cal L}^g(h)\right\| = {\cal L}^g_\coll(h) {\cal L}^g_\hc(h) \left({\cal L}^g_\qd(h)\right)^{\frac{N^{\max}_\qd}{N^g_\qd}}.
\label{eq:grasping.likelihood.norm}
\end{equation}
It is this normalised likelihood $\left\|{\cal L}^g\right\|$ that is used to rank all the generated grasps across all the grasp types during selection steps. After simulated annealing has yielded a ranked list of optimised grasp poses, they are checked for reachability given other objects in the workspace, and unreachable poses are pruned. 

\subsection{Grasp Execution}
The remaining best scoring hand pose $h*$ is then used to generate a reach to grasp trajectory. Since the hand is underactuated this consists of the wrist pose trajectory, and the motor signal trajectory. This is the command sequence that is executed on the robot.

%% file: inputTex/results.tex
\begin{figure*}
\begin{center}
\newcommand{\tw}{0.15\linewidth}
 \includegraphics[width=\tw]{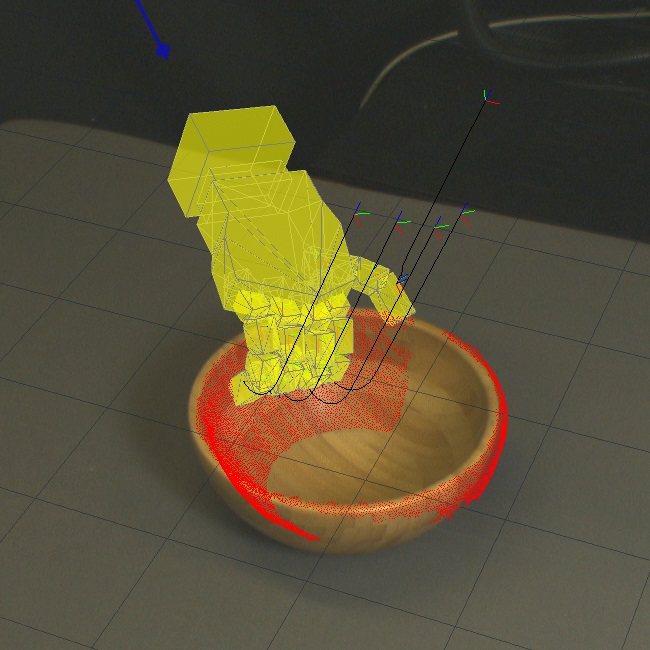} \hspace{-6pt}
 \includegraphics[width=\tw]{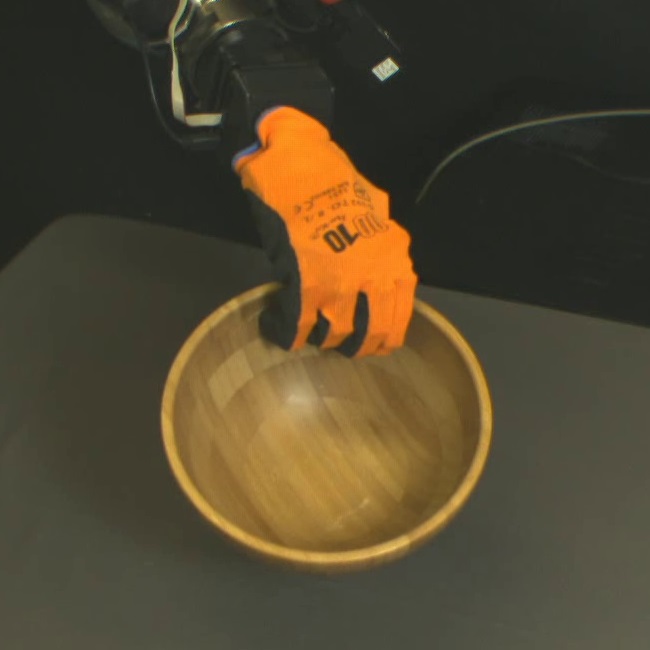}
 \includegraphics[width=\tw]{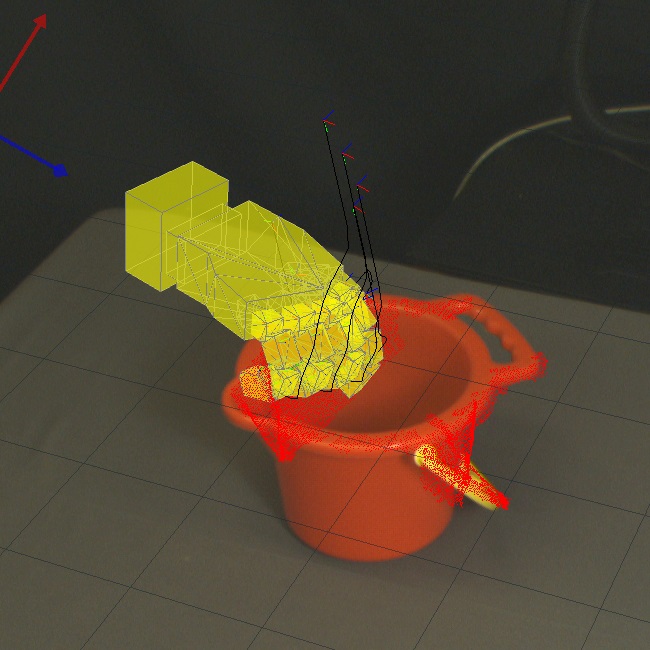} \hspace{-6pt}
 \includegraphics[width=\tw]{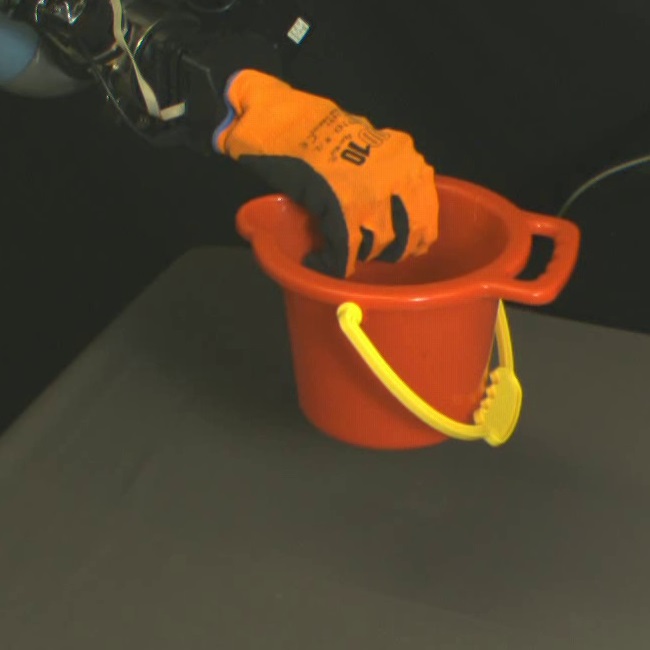}
  \includegraphics[width=\tw]{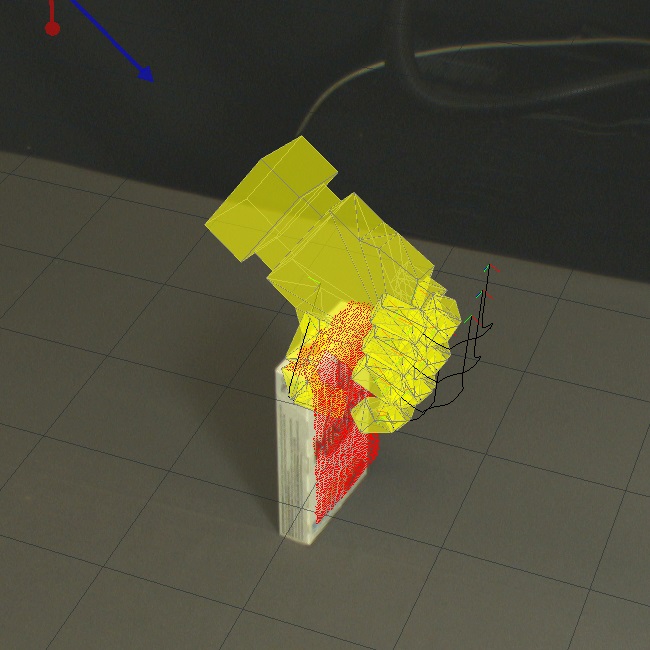} \hspace{-6pt}
 \includegraphics[width=\tw]{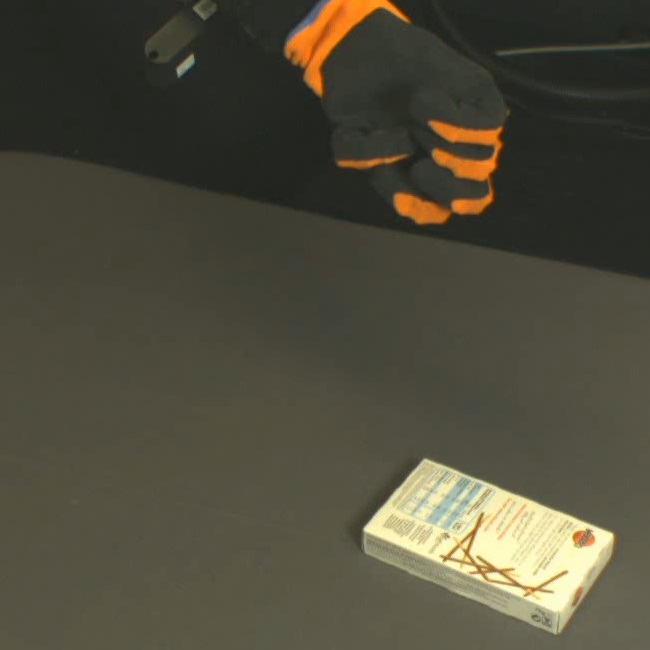}\\ [1ex]
  \includegraphics[width=\tw]{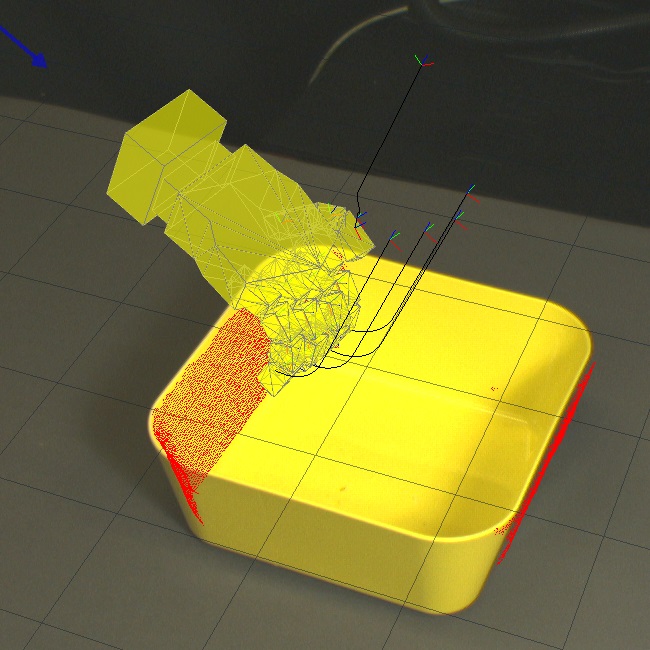} \hspace{-6pt}
 \includegraphics[width=\tw]{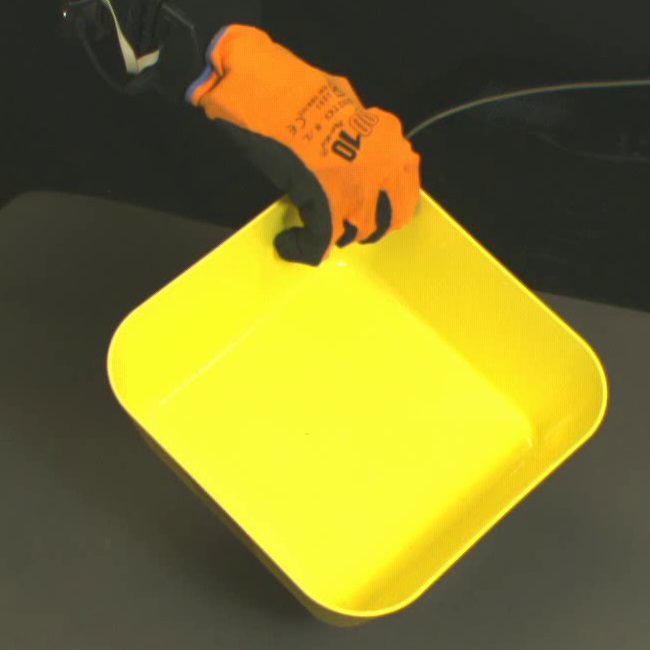}
  \includegraphics[width=\tw]{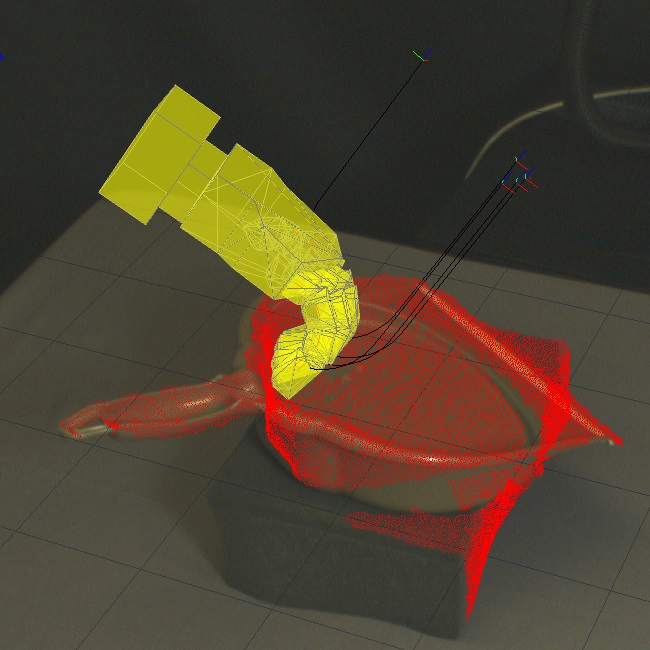} \hspace{-6pt}
 \includegraphics[width=\tw]{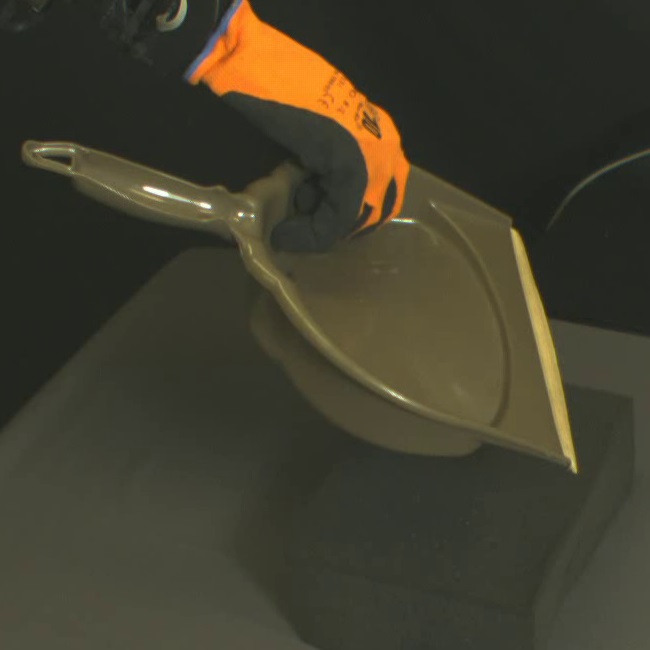}
  \includegraphics[width=\tw]{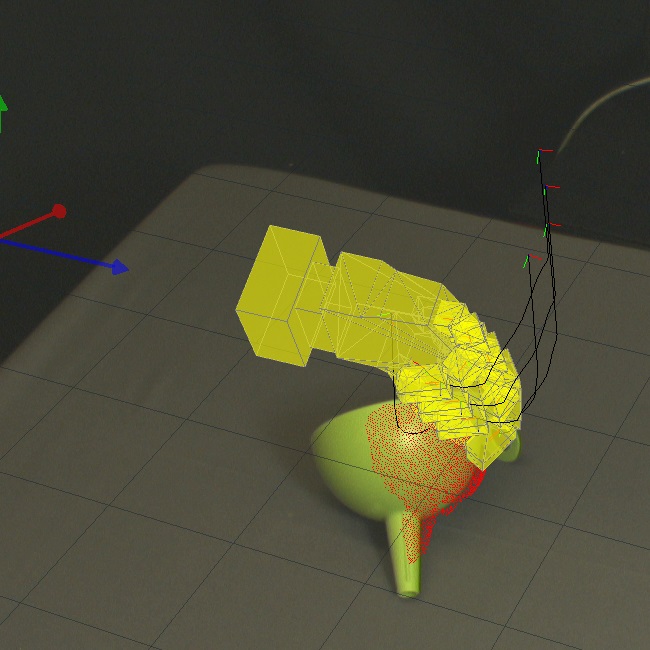} \hspace{-6pt}
 \includegraphics[width=\tw]{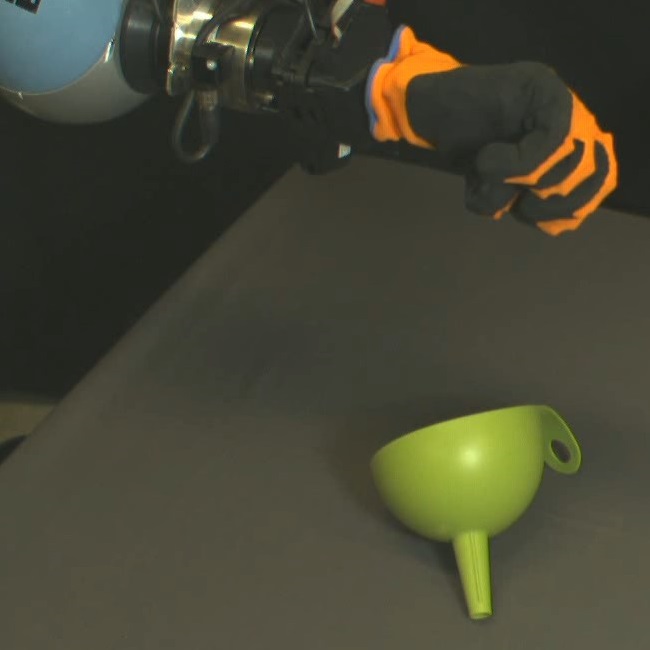}\\ [1ex]
  \includegraphics[width=\tw]{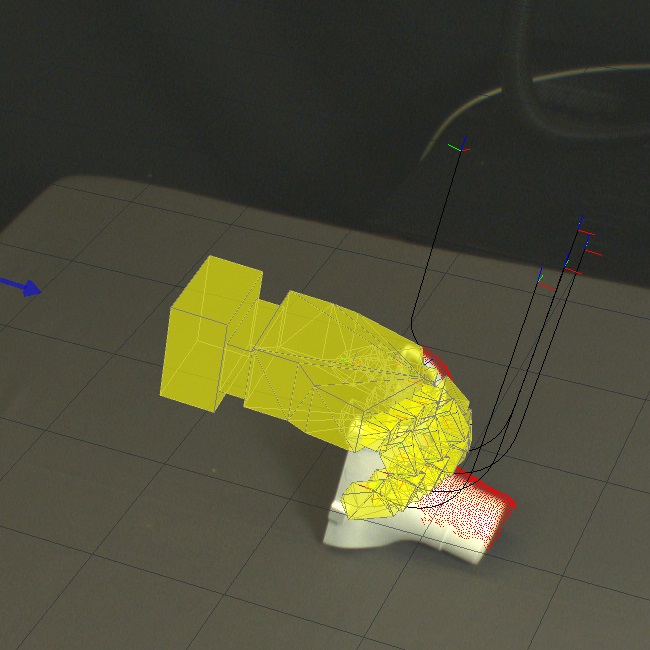} \hspace{-6pt}
 \includegraphics[width=\tw]{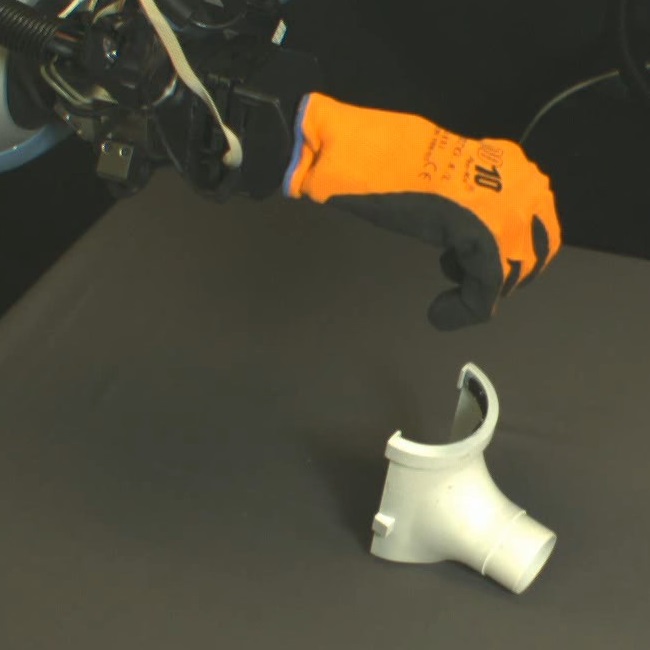}
  \includegraphics[width=\tw]{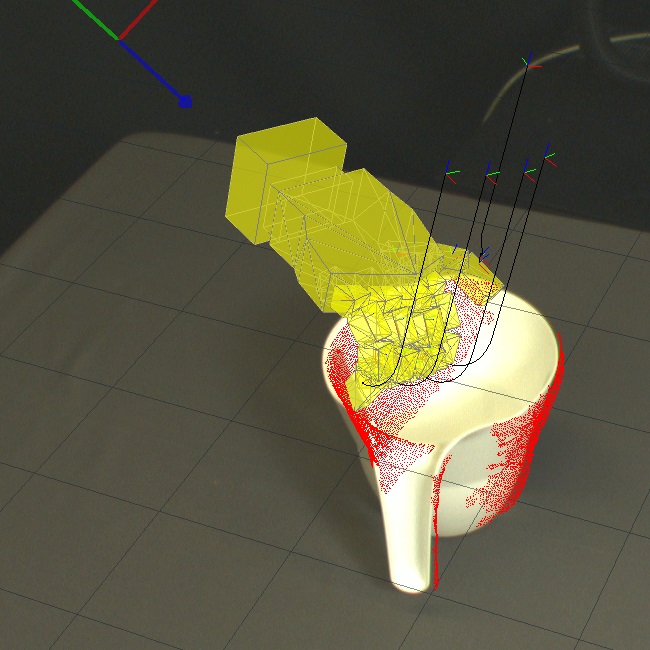} \hspace{-6pt}
 \includegraphics[width=\tw]{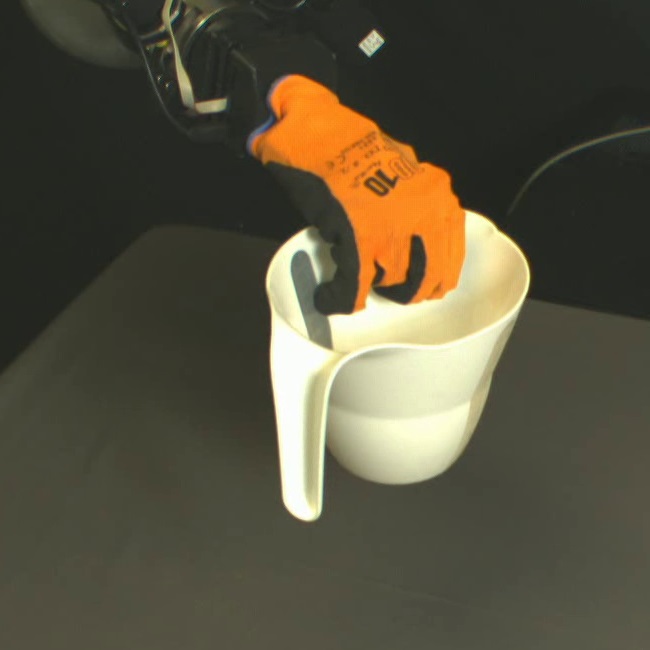}
  \includegraphics[width=\tw]{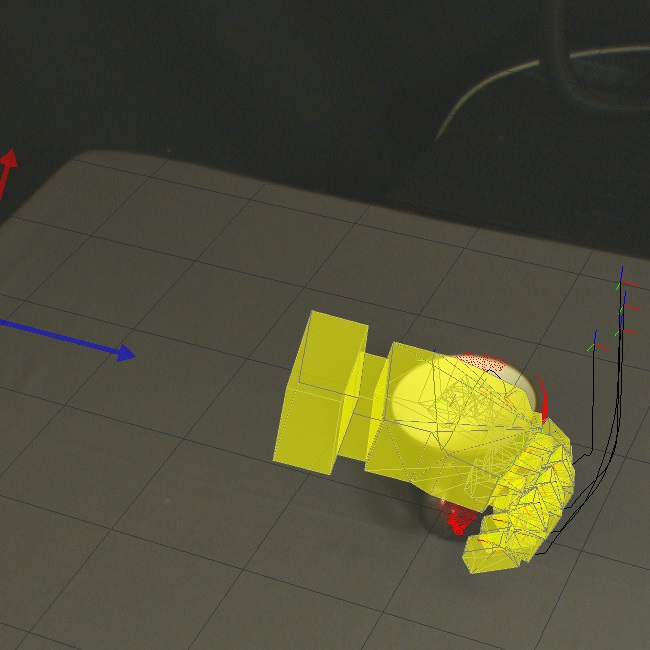} \hspace{-6pt}
 \includegraphics[width=\tw]{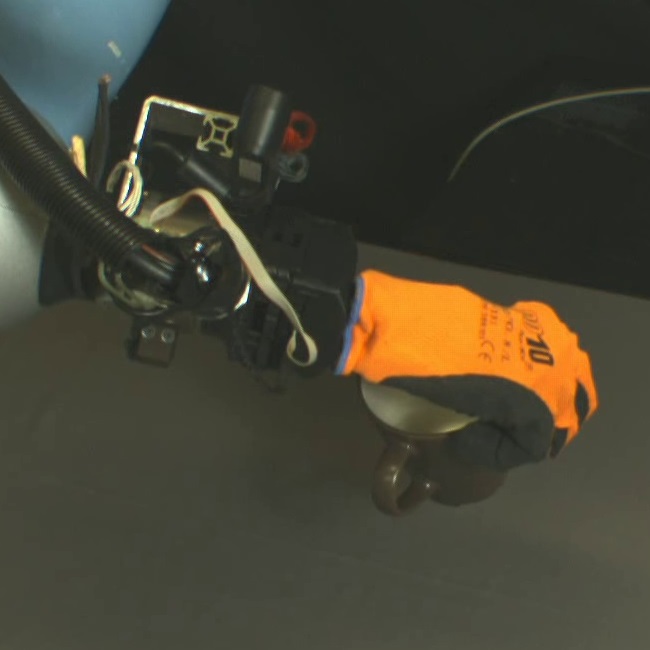}\\ [1ex]
  \includegraphics[width=\tw]{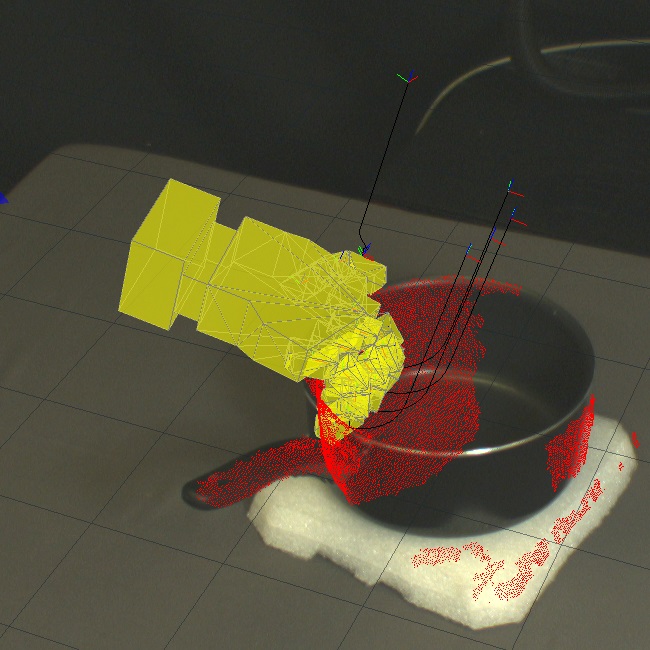} \hspace{-6pt}
 \includegraphics[width=\tw]{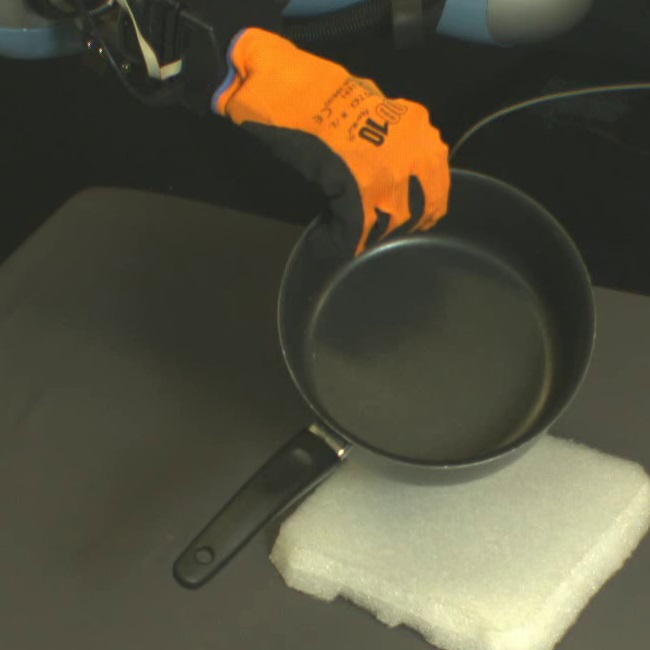}
  \includegraphics[width=\tw]{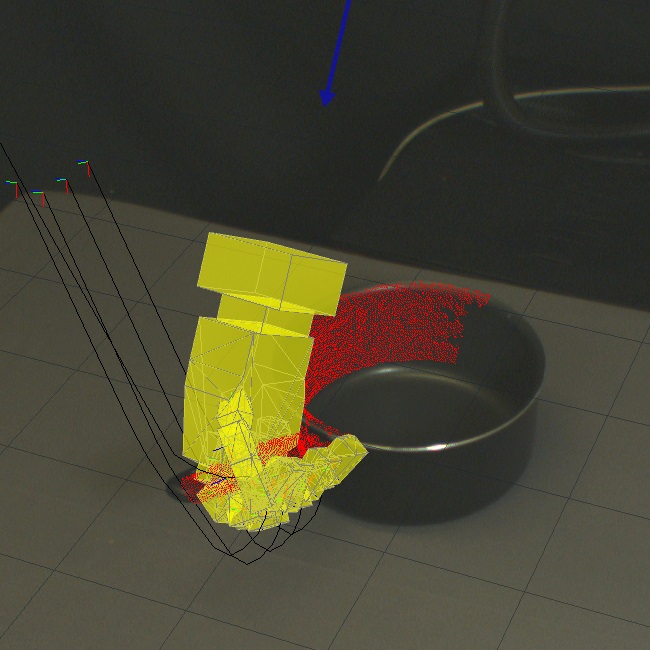} \hspace{-6pt}
 \includegraphics[width=\tw]{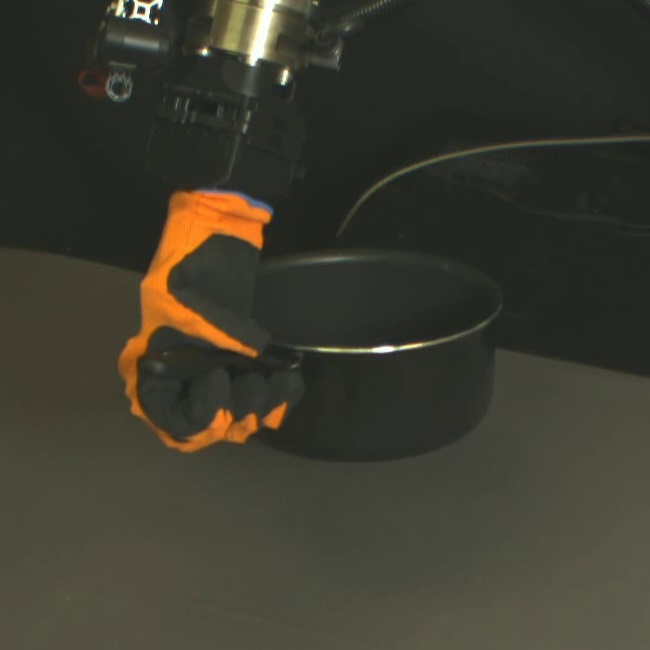} 
  \includegraphics[width=\tw]{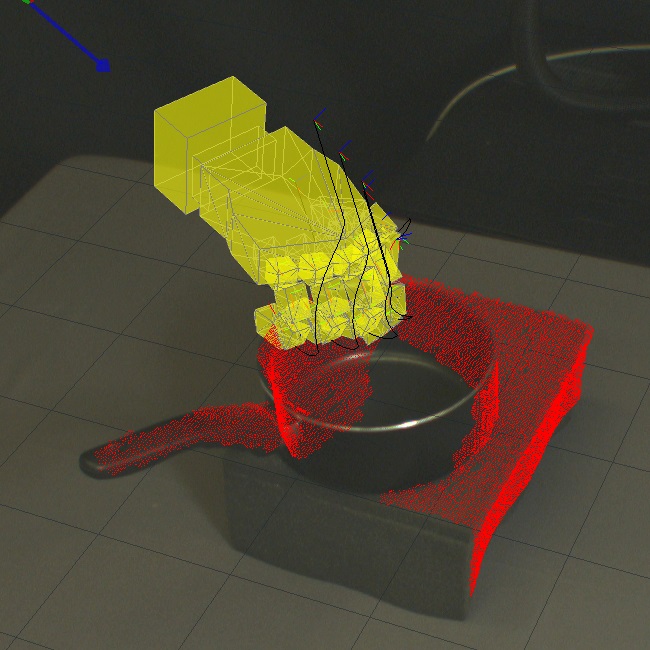} \hspace{-6pt}
 \includegraphics[width=\tw]{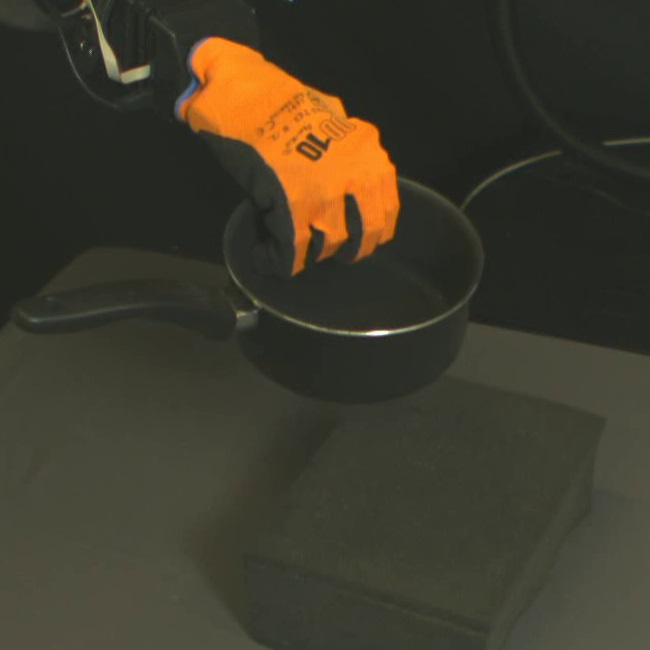}\\ [1ex]
  \includegraphics[width=\tw]{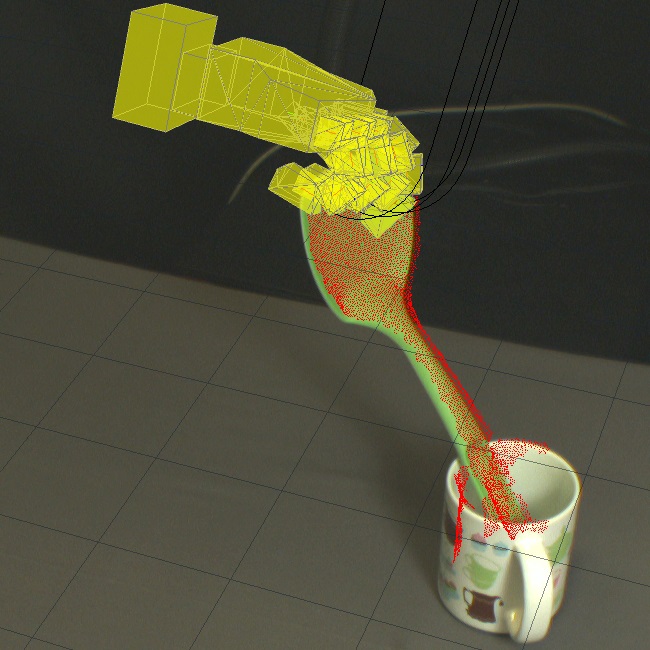} \hspace{-6pt}
 \includegraphics[width=\tw]{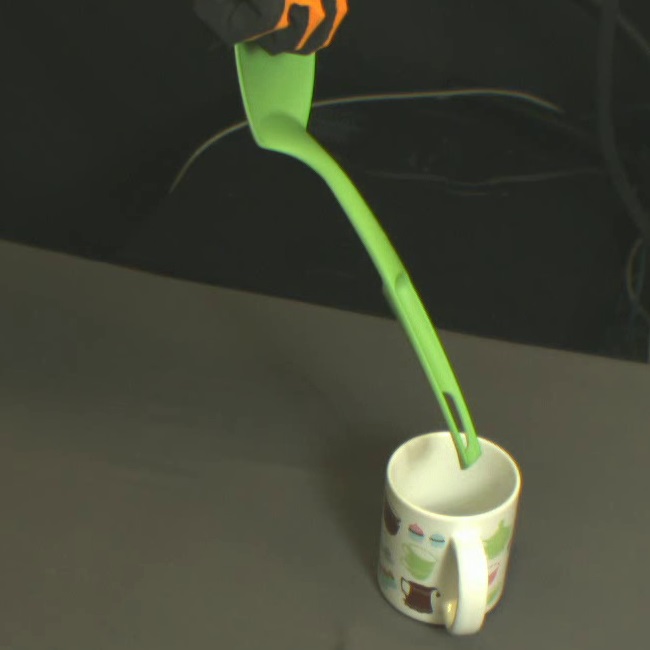}
  \includegraphics[width=\tw]{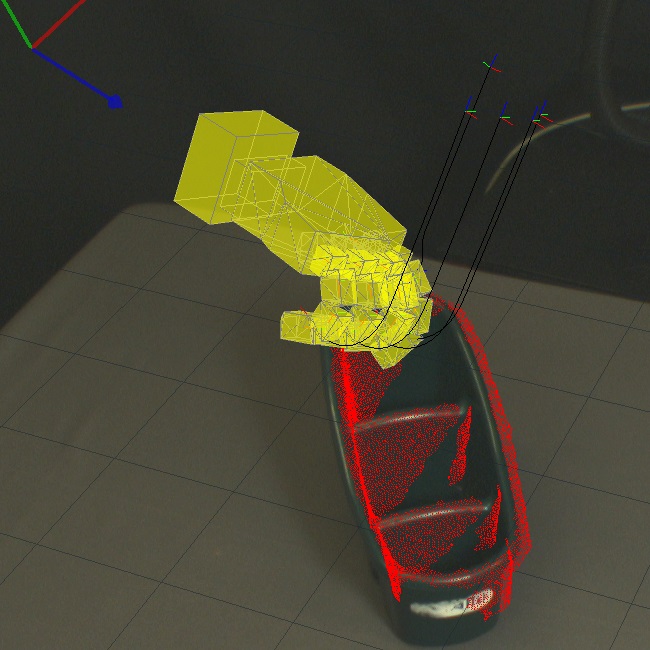} \hspace{-6pt}
 \includegraphics[width=\tw]{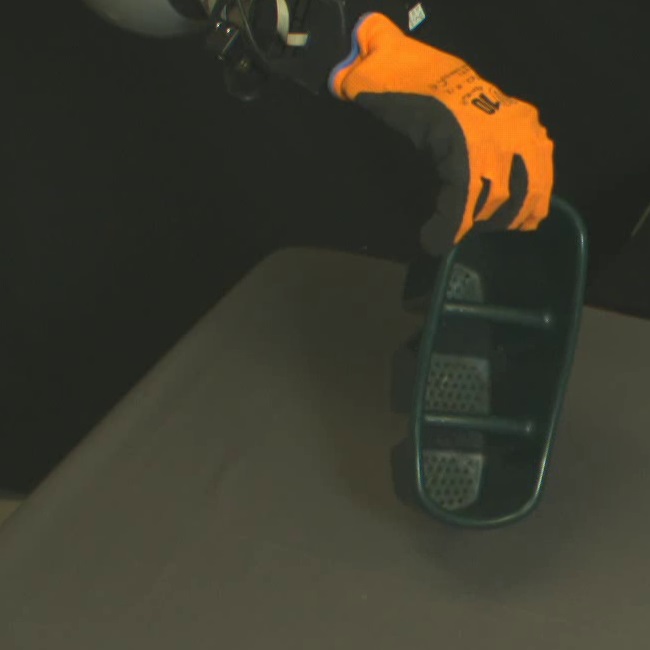}
  \includegraphics[width=\tw]{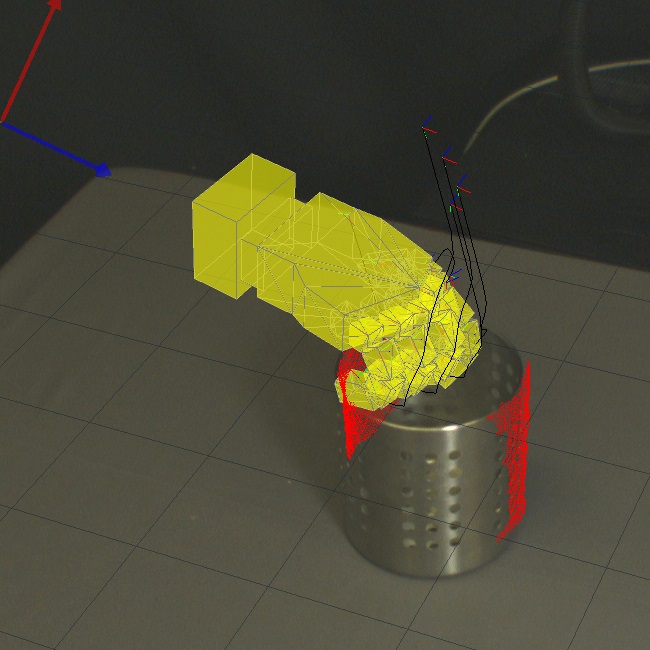} \hspace{-6pt}
 \includegraphics[width=\tw]{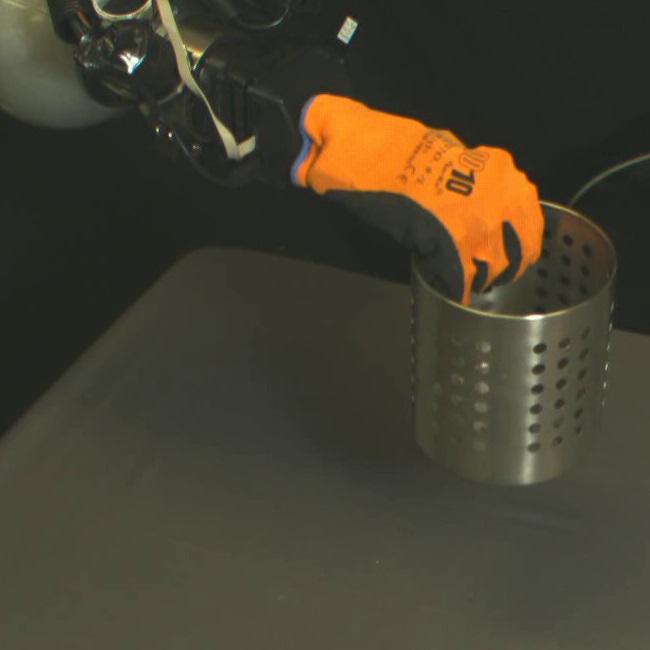}
 \caption{The fifteen test grasps. Each one has a pair of images. The predicted equilibrium grasp state is shown on the left of each pair, and the actual grasp on the right. Counting from top left it can be seen that grasps 3,6, and 7 failed. All other grasps succeeded.}
 \label{fig:test}
 \end{center}
\end{figure*}

The experiments were conducted as follows. Training consisted of nine example grasps, executed in simulation, with a human in control. These nine grasps were grouped into three grasp types (rim, pinch, and handle). The rim and pinch grasp types were trained on the colander object, and the handle grasp type was demonstrated on the saucepan. During testing an object was placed on the table. Every grasp type was compared automatically, and one selected for execution according to the methods described above. The models of the test objects consisted of a point cloud taken from just one view. Thus reconstructions were partial, typically less that 25\% of the object's surface area. No test objects had been seen previously by the robot, and it can be seen from Fig.~\ref{fig:test}. Fifteen test objects were presented, and 12 of the 15 test grasps succeeded, giving a generalisation success rate of 80\%. While the difference is not statistically significant, this is slightly higher than the 77.7\% success rate we recorded on a larger test set for a fully actuated hand also working from a single view of an object \cite{kopicki-detry-wyatt-etal-ijrr-2015}.

\begin{figure}
\begin{center}
 \includegraphics[width=0.9\columnwidth]{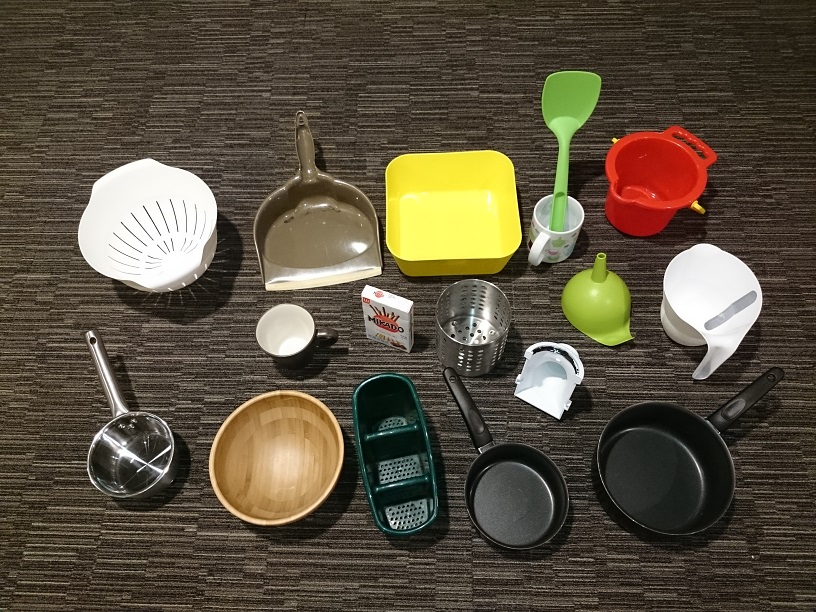}
 \caption{The two training objects are on the far left. The testing objects on the right. 12 from 15 test grasps on novel objects were successful.}
 \label{fig:training}
 \end{center}
\end{figure}

Finally, it is worth noting the similiarity between the location of the actual and targetted final grasp states. In a majority of cases the grasp involved interactions with the object, moving it to the stable grasp pose. This is a natural property of the hand, but it might have been supposed that the learning method would not be robust to such interactions in terms of the accuracy of the grasp.

%% file: soft-grasp-arxiv2016.bbl
\begin{thebibliography}{10}
\providecommand{\url}[1]{#1}
\csname url@rmstyle\endcsname
\providecommand{\newblock}{\relax}
\providecommand{\bibinfo}[2]{#2}
\providecommand\BIBentrySTDinterwordspacing{\spaceskip=0pt\relax}
\providecommand\BIBentryALTinterwordstretchfactor{4}
\providecommand\BIBentryALTinterwordspacing{\spaceskip=\fontdimen2\font plus
\BIBentryALTinterwordstretchfactor\fontdimen3\font minus
  \fontdimen4\font\relax}
\providecommand\BIBforeignlanguage[2]{{%
\expandafter\ifx\csname l@#1\endcsname\relax
\typeout{** WARNING: IEEEtran.bst: No hyphenation pattern has been}%
\typeout{** loaded for the language `#1'. Using the pattern for}%
\typeout{** the default language instead.}%
\else
\language=\csname l@#1\endcsname
\fi
#2}}

\bibitem{saxena2008b}
A.~Saxena, L.~Wong, and A.~Ng, ``Learning grasp strategies with partial shape
  information,'' in \emph{Proceedings of AAAI}.\hskip 1em plus 0.5em minus
  0.4em\relax AAAI, 2008, pp. 1491--1494.

\bibitem{detry2013a}
R.~Detry, C.~H. Ek, M.~Madry, and D.~Kragic, ``Learning a dictionary of
  prototypical grasp-predicting parts from grasping experience,'' in
  \emph{International Conference on Robotics and Automation}.\hskip 1em plus
  0.5em minus 0.4em\relax IEEE, 2013, pp. 601--608.

\bibitem{herzog2014a}
A.~Herzog, P.~Pastor, M.~Kalakrishnan, L.~Righetti, J.~Bohg, T.~Asfour, and
  S.~Schaal, ``Learning of grasp selection based on shape-templates,''
  \emph{Autonomous Robots}, vol.~36, no. 1-2, pp. 51--65, 2014.

\bibitem{kroemer2012a}
O.~Kroemer, E.~Ugur, E.~Oztop, and J.~Peters, ``A kernel-based approach to
  direct action perception,'' in \emph{{IEEE} International Conference on
  Robotics and Automation}.\hskip 1em plus 0.5em minus 0.4em\relax IEEE, 2012,
  pp. 2605--2610.

\bibitem{ben2012generalization}
H.~Ben~Amor, O.~Kroemer, U.~Hillenbrand, G.~Neumann, and J.~Peters,
  ``Generalization of human grasping for multi-fingered robot hands,'' in
  \emph{International Conference on Intelligent Robots and Systems}.\hskip 1em
  plus 0.5em minus 0.4em\relax IEEE, 2012, pp. 2043--2050.

\bibitem{hillenbrand2012transferring}
U.~Hillenbrand and M.~Roa, ``Transferring functional grasps through contact
  warping and local replanning,'' in \emph{IEEE/RSJ International Conference on
  Robotics and Systems}.\hskip 1em plus 0.5em minus 0.4em\relax IEEE, 2012, pp.
  2963--2970.

\bibitem{Catalano2014Adaptive}
M.~G. Catalano, G.~Grioli, E.~Farnioli, A.~Serio, C.~Piazza, and A.~Bicchi,
  ``Adaptive synergies for the design and control of the {Pisa/IIT}
  {SoftHand},'' \emph{The International Journal of Robotics Research}, vol.~33,
  no.~5, pp. 768--782, 2014.

\bibitem{Dollar2010Highly}
A.~M. Dollar and R.~D. Howe, ``The highly adaptive {SDM} hand: Design and
  performance evaluation,'' \emph{The International Journal of Robotics
  Research}, vol.~29, no.~5, pp. 585--597, 2010.

\bibitem{Eppner2015Planning}
C.~Eppner and O.~Brock, ``Planning grasp strategies that exploit environmental
  constraints,'' in \emph{Proceedings of the IEEE International Conference on
  Robotics and Automation}, 2015.

\bibitem{Bonilla2015Grasp}
M.~Bonilla, D.~Resaco, M.~Gabiccini, and A.~Bicchi, ``Grasp planning with soft
  hands using bounding box object decomposition,'' in \emph{In Proceedings of
  the IEEE International Conference of Intelligent Robots and Systems}, 2015.

\bibitem{silverman1986a}
B.~W. Silverman, \emph{Density Estimation for Statistics and Data
  Analysis}.\hskip 1em plus 0.5em minus 0.4em\relax {Chapman \& Hall/CRC},
  1986.

\bibitem{fisher1953a}
R.~A. Fisher, ``Dispersion on a sphere,'' in \emph{Proc. Roy. Soc. London Ser.
  A.}, vol. 217, no. 1130.\hskip 1em plus 0.5em minus 0.4em\relax Royal
  Society, 1953, pp. 295--305.

\bibitem{sudderth2006a}
E.~B. Sudderth, ``Graphical models for visual object recognition and
  tracking,'' Ph.D. dissertation, MIT, Cambridge, MA, 2006.

\bibitem{kanatani2005statistical}
K.~Kanatani, \emph{Statistical optimization for geometric computation: theory
  and practice}.\hskip 1em plus 0.5em minus 0.4em\relax Courier Dover
  Publications, 2005.

\bibitem{spivak1979comprehensive}
M.~Spivak, \emph{A comprehensive introduction to differential geometry}.\hskip
  1em plus 0.5em minus 0.4em\relax Publish or Perish Berkeley, 1999, vol.~1.

\bibitem{kirkpatrick83optimizationby}
S.~Kirkpatrick, C.~D. Gelatt, and M.~P. Vecchi, ``Optimization by simulated
  annealing,'' \emph{Science}, vol. 220, no. 4598, pp. 671--680, 1983.

\bibitem{kopicki-detry-wyatt-etal-ijrr-2015}
M.~Kopicki, R.~Detry, M.~Adjigble, A.~Leonardis, and J.~L. Wyatt, ``One shot
  learning and generation of dexterous grasps for novel objects,''
  \emph{International Journal of Robotics Research}, 2015.

\end{thebibliography}
